\documentclass[11pt,nonatbib]{elsarticle}
\usepackage{amsmath,amssymb,amsthm,mathtools}
\usepackage{mathrsfs}
\usepackage[cal=euler]{mathalpha}
\usepackage{dsfont}
\usepackage{xfrac}
\usepackage{graphicx}
\usepackage{rotating}
\usepackage{array}
\usepackage{makecell}
\usepackage{booktabs}
\usepackage{tabularx}
\usepackage{siunitx}
\usepackage{threeparttable}
\usepackage{multirow}
\usepackage{float}
\usepackage{subfigure} 

\usepackage{tikz}
\usepackage{pgfplotstable}
\usetikzlibrary{
  calc,
  fit,
  decorations.pathreplacing,
  chains,arrows.meta
}
\usepgfplotslibrary{groupplots,fillbetween}
\usepackage{enumerate}
\usepackage{calc}
\usepackage{comment}
\usepackage{xcolor}
\usepackage{soul}
\usepackage{ifthen}
\usepackage{xspace}
\usepackage{listings,mlmodern}

\usepackage[
  backend=biber,
  style=authoryear
]{biblatex}
\usepackage{hyperref}

\hypersetup{
  pdfauthor={BGLSC},
  pdftitle={NTML},
  colorlinks=true,
  linkcolor=blue,
  urlcolor=orange,
  citecolor=blue,
  linktocpage=true,
  pdfstartpage=1,
  pdfstartview=FitV,
  breaklinks=true,
  pdfpagemode=UseNone,
  bookmarksopen=true,
  bookmarksnumbered=true
}

\DeclareCiteCommand{\parencite}[\mkbibparens]
  {\usebibmacro{prenote}}
  {%
    \usebibmacro{citeindex}%
    \printtext[bibhyperref]{\usebibmacro{cite}}%
  }
  {\multicitedelim}
  {\usebibmacro{postnote}}

\let\cite\parencite

\newcommand{\NT}{\mathds{N}\mathcal{T}}
\newcommand{\ND}{\mathds{N}\mathcal{D}}
\newcommand{\R}{\mathds{R}}

\newcommand{\bt}{\boldsymbol{t}}
\newcommand{\bw}{\boldsymbol{w}}

\newcommand{\bh}{\boldsymbol{h}}
\newcommand{\bk}{\boldsymbol{k}}
\newcommand{\bq}{\boldsymbol{q}}
\newcommand{\btheta}{\boldsymbol{\theta}}

\newcommand{\btau}{\boldsymbol{\tau}}

\newlength{\dunder}
\theoremstyle{definition}

\begin{document}

\begin{frontmatter}

\title{Testing Transformer Learnability\\
on the Arithmetic Sequence of Rooted Trees}

\author[aff1,aff3]{Alessandro Breccia\corref{cor1}}
\ead{alessandro.breccia.25@ucl.ac.uk}

\author[aff1]{Federica Gerace}
\author[aff2]{Marco Lippi}
\author[aff1]{Gabriele Sicuro}
\author[aff1]{Pierluigi Contucci}

\cortext[cor1]{Corresponding author}

\address[aff1]{%
  Department of Mathematics,
  University of Bologna,
  Bologna, Italy
}

\address[aff2]{%
  Department of Information Engineering,
  University of Florence,
  Florence, Italy
}

\address[aff3]{%
  Gatsby Computational Neuroscience Unit,
  University College London,
  London, UK
}

\begin{abstract}
We study whether a transformer network can learn the deterministic sequence of trees generated by the iterated prime factorization of the natural numbers. Each integer is mapped into a rooted planar tree and the resulting sequence $\NT$ defines an arithmetic text with measurable statistical structure. A transformer network (the GPT-2 architecture) is trained from scratch on the first $10^{11}$ elements and evaluated on next-token prediction and masked-token prediction tasks, with a Hidden Markov Model as baseline and a scaling analysis over context window, dataset size, vocabulary size and model size. The model reaches a word accuracy of about $0.4$, well above the baseline, and its performance remains stable on test blocks located at $10^{13}$--$10^{15}$, far beyond the training interval. Moreover, the likelihood assigned by the trained model separates the arithmetic text from two controls: synthetic token sequences that preserve the marginal token frequencies but not their order, with a separation that widens with the context, and sequences containing more than three consecutive square-free integers, a configuration that arithmetic forbids. 
These results indicate that the transformer captures regularities of the arithmetic text
that go beyond its frequency profile.
\end{abstract}

\end{frontmatter}

\section{Introduction}

Large Language Models (LLM) have shown a remarkable ability to capture structure
in human language, biological sequences, and symbolic domains such as
mathematics. Yet the emergence of this ability, i.e., the capacity to extract
latent rules from raw sequences, remains mostly an empirical observation
rather than a theoretically understood mechanism. A natural strategy to
probe its foundations is to train a model on a corpus whose generative rule
is known exactly: when the data-generating law is available, the question of
what the model has learned becomes an experimental one, since predictions
can be measured against the structure that actually produces the text. This
strategy has recently become an important experimental paradigm.
Markovian settings have been employed to study the emergence of in-context
learning, induction heads, and associative memory by training transformers
on sequences generated from random transition kernels
\cite{edelman2024evolution,nichani2025understanding}; synthetic hierarchical
languages generated by artificial grammars have been used to investigate how
transformers acquire hierarchical representations and their scaling
behavior \cite{allen2023physics,cagnetta2024towards}; and recursive
synthetic languages have served to analyze the emergence of compositional
reasoning and the learning of context-free grammatical structures
\cite{qin2024sometimes,schulz2025unraveling}. In all these settings,
however, the data-generating process is itself an object of design: the
transition kernel, grammar, or recursive rules are chosen by the
experimenter, often sampled from a parametrized family of synthetic models.

In this work we consider a corpus whose ``grammar'' is not designed but
inherited from arithmetic. Every integer, through its prime factorization
iterated to the exponents, is mapped to a rooted planar tree (see
Eq.~\ref{eq:fig} below) encoding its multiplicative and exponential prime
structure, or equivalently to a Dyck word, namely a balanced binary string.
The resulting infinite sequence, denoted by $\NT$, is a deterministic text
generated by a unique canonical rule. Some of its number-theoretical and statistical properties have been
studied in a series of works
\cite{conti2023,Conti2024,CCI2,Modena2025}: the dictionary of distinct Dyck
words grows sublinearly, the rank--frequency distribution is scale-stable
and departs from Zipf's law, the text is measurably oriented under
word-order reversal, and correlation analysis reveals genuine long-range
dependencies with a diffusive-to-superdiffusive crossover. 

{
This double reading --- a deterministic arithmetic law on one side, a symbolic corpus with measurable statistics on the other --- places the present work in a line of long-standing problems in statistical physics, where arithmetic quantities appear as free energies and phase transitions: the
number-theoretical spin chain \cite{Knauf1993} and the Hecke-algebra system of \textcite{BostConnes1995} are two instances. What distinguishes $\NT$ is that
no Hamiltonian is imposed from outside: the ensemble is the one arithmetic
itself generates, and the observables --- dictionary growth, rank--frequency
hierarchy, entropy, correlation exponents \cite{Modena2025} --- are measured
directly on the sequence. The learning problem inherits the same formalism,
the generative law of Eq.~\eqref{eq:soft} being a Boltzmann weight over the
dictionary at inverse temperature $\beta$, and the questions we ask about the
descent of the loss and its scaling are those posed for learning machines in
this tradition \cite{cagnetta2024towards,cagnetta2025scaling,rende2024distributional}.
Corpus, statistical description and learning machine are three levels of a
single ensemble-theoretic account.
}

As a controlled testbed for learnability, $\NT$ differs from designed
synthetic corpora in three respects. First, its statistical regularities
are not engineered to facilitate learning, but emerge from arithmetic
constraints such as unique factorization and the recursive structure of
exponents. Second, unlike settings that generate multiple independent
sequences sampled from a family of stochastic models, there is a single
canonical arithmetic trajectory: there is only one $\NT$. Finally, every
symbol and every constraint admits an explicit mathematical
interpretation, so that the learned behavior can be evaluated against
independently defined number-theoretic properties such as primality,
square-freeness, and provably impossible configurations of words. The
question addressed here is therefore not whether a transformer can infer an
unknown synthetic grammar, but to what extent it can learn the statistical
structure of a deterministic mathematical sequence whose generating law is
known exactly. We trained the transformer on the sequence and evaluated its sensitivity to the sequential organization of the corpus and to an independently specified arithmetic constraint.

The representation itself also distinguishes this work from previous
machine learning approaches to arithmetic sequences. Neural networks
applied to prime-related predicates \cite{He2021} showed limited ability to
reproduce their distribution, and theoretical arguments based on
Kolmogorov complexity \cite{KolpakovRocke2023,KolpakovRocke2024} suggest
that the prime indicator function is not compressible within standard
statistical frameworks. Related work on modular classification of integers
\cite{Wu2023} shows that high accuracy is obtained only when externally
provided arithmetic features are incorporated into the data representation.
Attempts to predict prime gaps or infer prime ratios using sequential or
dense architectures \cite{Pylov2023,Blake2023} demonstrate local predictive
ability over restricted ranges, but accuracy degrades as size increases and
no structural inference emerges. Early neural factorization experiments
\cite{Jansen2005} concluded that numerical encodings behave as noise beyond
superficial correlations, a finding confirmed with modern models
\cite{NeneUludag2022}, where degradation with bit length strongly suggests
that structural information is absent from raw inputs. Diffusion-based
refinement techniques \cite{Freivalds2023} have produced improved numerical
results on limited ranges, but scaling issues remain. Machine learning
methods have also been used to approximate the M\"obius function and
related arithmetic predicates \cite{QinYe2024,LeeKim2024}, often achieving
high accuracy, but only when the input includes explicit structured
information such as modular reductions or sparse encodings. Most of these
studies thus operate on representations that expose only partial aspects of
the multiplicative structure of the integers, whether via numerical
encoding, handcrafted features, or modular reductions. Here, instead, the
integers enter the model through their complete multiplicative tree
structure, encoded as Dyck words: no external arithmetic descriptor is
supplied, and whatever structure is learned must be extracted from the raw
deterministic text.

The natural question that follows is whether this arithmetic language is
learnable. More precisely, \emph{can a transformer exposed to the first
portion of $\NT$ infer the continuation of the sequence and fill in the
gaps?} We address this question by training a transformer network (the
GPT-2 architecture, taken as canonical) from scratch on the sequence of
Dyck words associated with the first $10^{11}$ integers under the two
canonical self-supervised tasks of Next-Token Prediction (NTP) and Masked
Language Modelling (MLM). Before training, the Dyck-word sequence is converted into a sequence of tokens using Byte-Pair Encoding (BPE). Because BPE tokens need not coincide with Dyck-word boundaries, the model operates at the token level, whereas some evaluation metrics are computed on the reconstructed Dyck-word sequence. We quantify the performance both at the level of
generic Dyck words and on specific words of arithmetic interest, such as
primes, against a hidden Markov model baseline; we study how the results depend
on context window, dataset size, vocabulary size and model size; we test
the stability of the predictions on intervals of integers far beyond the
training range; and we ask whether the likelihood assigned by the trained model discriminates the arithmetic text from synthetic token sequences preserving the empirical marginal BPE-token distribution in expectation but lacking the sequential organization of the original corpus, and whether it penalizes a configuration that arithmetic forbids, namely more than three consecutive square-free integers.

These experiments shed light on two distinct yet complementary aspects. On
the number-theoretic side, they measure the degree to which a neural
architecture designed for human language can predict the tree structure of
an integer from the preceding ones. On the machine-learning side, they use
a mathematically grounded corpus to probe what a transformer actually
recovers of a known generating law. We note that prime factorization is
also a computational problem of independent interest, whose difficulty at
scale makes any learning-based approach potentially attractive; questions
regarding efficiency and comparison with existing factorization algorithms
are, however, beyond the scope of this work and are left for future
investigation.

The remainder of the paper is organized as follows. Section~\ref{sec:data}
introduces the dataset and notation, Section~\ref{sec:model} describes the
model architecture, Section~\ref{sec:tasks} defines the training tasks, and
Section~\ref{sec:experiments} reports the experimental results for NTP and
MLM together with baseline comparisons and the scaling analysis. We also assess whether the trained model has
internalized known structural constraints of $\NT$, and
Section~\ref{sec:conclusions} concludes.

\section{The Dataset $\mathds N\mathcal T_n$}\label{sec:data}
To produce our dataset, we proceed as follows. Let $n\in\mathds{N}$ be a natural number and $\omega(n)$ the cardinality of the set of its prime factors. Let us consider the following recursive definition of $\tau(n)$ \cite{conti2023}:
\begin{itemize}
    \item if $n=1$, then $\tau(n)=\{1\colon 1\}$;
    \item if $n>1$, let $n=\prod_{k=1}^{\omega(n)} p_k^{a_k}$ be its factorization in $\omega(n)$ distinct primes, $a_k\in\mathds{N}$, labeled in such a way that if $k<k'\Rightarrow p_k<p_{k'}$; then $\tau(n)=\{p_k\colon \tau(a_k)\}_{k=1}^{\omega(n)}$.
\end{itemize}
Such a representation of a natural number has also been considered, from a different point of view, by \textcite{devlin2014} --- who called it \textit{prime tower factorization} --- and it is naturally associated to a tree \cite{Childress2021,Iudelevich2022}, in which a root, associated to $n$, is linked to $\omega(n)$ ``children'', each corresponding to one of the prime factors of $n$. Given the child $p_k$, we iterate the construction associating to it $\omega(a_k)$ children, one for each prime factor of its exponent's prime decomposition. The construction terminates when an exponent equal to one is reached. For example, for $n=55340232221128654848$, we would naturally construct the representation
\begin{multline}n=2^{64}\cdot 3=2^{2^{2\cdot 3}}\cdot 3\mapsto\\ \tau(n)=\{2:\{2:\{\{2:\{1:1\}\},\{3:\{1:1\}\}\}\},3:\{1:1\}\}\mapsto  \begin{tikzpicture}[baseline={([yshift=-.5ex]current bounding box.center)}]
\node[inner sep=1pt,circle,draw,thick,fill=gray] (0) at (0,0) {\small $n$};
\node[inner sep=1pt,circle,draw,thick] (1) at (-0.25,0.5) {\small $2$};
\node[inner sep=1pt,circle,draw,thick] (2) at (0.25,0.5) {\small $3$};
\node[inner sep=1pt,circle,draw,thick] (12) at (-0.25,1) {\small $2$};
\node[inner sep=1pt,circle,draw,thick] (121) at (-0.5,1.5) {\small $2$};
\node[inner sep=1pt,circle,draw,thick] (122) at (0,1.5) {\small $3$};
\draw[thick] (122) -- (12) -- (1) -- (0) -- (2);
\draw[thick] (121) -- (12);
\end{tikzpicture}\label{eq:fig}\end{multline}
The natural numbers up to $16$ can be represented as
\[\resizebox{\linewidth}{!}{\begin{tikzpicture}
\node[inner sep=1pt,circle,draw,thick,fill=gray] (0) at (0,0) {\small $1$};
\end{tikzpicture}\quad\begin{tikzpicture}
\node[inner sep=1pt,circle,draw,thick,fill=gray] (0) at (0,0) {\small $2$};
\node[inner sep=1pt,circle,draw,thick] (1) at (0,0.5) {\small $2$};
\draw[thick] (0) -- (1);
\end{tikzpicture}\quad\begin{tikzpicture}
\node[inner sep=1pt,circle,draw,thick,fill=gray] (0) at (0,0) {\small $3$};
\node[inner sep=1pt,circle,draw,thick] (1) at (0,0.5) {\small $3$};
\draw[thick] (0) -- (1);
\end{tikzpicture}\quad\begin{tikzpicture}
\node[inner sep=1pt,circle,draw,thick,fill=gray] (0) at (0,0) {\small $4$};
\node[inner sep=1pt,circle,draw,thick] (1) at (0,0.5) {\small $2$};
\node[inner sep=1pt,circle,draw,thick] (2) at (0,1) {\small $2$};
\draw[thick] (0) -- (1) -- (2);
\end{tikzpicture}\quad\begin{tikzpicture}
\node[inner sep=1pt,circle,draw,thick,fill=gray] (0) at (0,0) {\small $5$};
\node[inner sep=1pt,circle,draw,thick] (1) at (0,0.5) {\small $5$};
\draw[thick] (0) -- (1);
\end{tikzpicture}\quad\begin{tikzpicture}
\node[inner sep=1pt,circle,draw,thick,fill=gray] (0) at (0,0) {\small $6$};
\node[inner sep=1pt,circle,draw,thick] (1) at (-0.25,0.5) {\small $2$};
\node[inner sep=1pt,circle,draw,thick] (2) at (0.25,0.5) {\small $3$};
\draw[thick] (2) -- (0) -- (1);
\end{tikzpicture}\quad\begin{tikzpicture}
\node[inner sep=1pt,circle,draw,thick,fill=gray] (0) at (0,0) {\small $7$};
\node[inner sep=1pt,circle,draw,thick] (1) at (0,0.5) {\small $7$};
\draw[thick] (0) -- (1);
\end{tikzpicture}\quad\begin{tikzpicture}
\node[inner sep=1pt,circle,draw,thick,fill=gray] (0) at (0,0) {\small $8$};
\node[inner sep=1pt,circle,draw,thick] (1) at (0,0.5) {\small $2$};
\node[inner sep=1pt,circle,draw,thick] (2) at (0,1) {\small $3$};
\draw[thick] (0) -- (1) -- (2);
\end{tikzpicture}\quad\begin{tikzpicture}
\node[inner sep=1pt,circle,draw,thick,fill=gray] (0) at (0,0) {\small $9$};
\node[inner sep=1pt,circle,draw,thick] (1) at (0,0.5) {\small $3$};
\node[inner sep=1pt,circle,draw,thick] (2) at (0,1) {\small $2$};
\draw[thick] (0) -- (1) -- (2);
\end{tikzpicture}\quad\begin{tikzpicture}
\node[inner sep=0pt,circle,draw,thick,fill=gray] (0) at (0,0) {\small $10$};
\node[inner sep=1pt,circle,draw,thick] (1) at (-0.25,0.5) {\small $2$};
\node[inner sep=1pt,circle,draw,thick] (2) at (0.25,0.5) {\small $5$};
\draw[thick] (2) -- (0) -- (1);
\end{tikzpicture}\quad\begin{tikzpicture}
\node[inner sep=0pt,circle,draw,thick,fill=gray] (0) at (0,0) {\small $11$};
\node[inner sep=0pt,circle,draw,thick] (1) at (0,0.5) {\small $11$};
\draw[thick] (0) -- (1);
\end{tikzpicture}\quad\begin{tikzpicture}
\node[inner sep=0pt,circle,draw,thick,fill=gray] (0) at (0,0) {\small $12$};
\node[inner sep=1pt,circle,draw,thick] (1) at (-0.25,0.5) {\small $2$};
\node[inner sep=1pt,circle,draw,thick] (11) at (-0.25,1) {\small $2$};
\node[inner sep=1pt,circle,draw,thick] (2) at (0.25,0.5) {\small $3$};
\draw[thick] (2) -- (0) -- (1) -- (11);
\end{tikzpicture}\quad\begin{tikzpicture}
\node[inner sep=0pt,circle,draw,thick,fill=gray] (0) at (0,0) {\small $13$};
\node[inner sep=0pt,circle,draw,thick] (1) at (0,0.5) {\small $13$};
\draw[thick] (0) -- (1);
\end{tikzpicture}\quad\begin{tikzpicture}
\node[inner sep=0pt,circle,draw,thick,fill=gray] (0) at (0,0) {\small $14$};
\node[inner sep=1pt,circle,draw,thick] (1) at (-0.25,0.5) {\small $2$};
\node[inner sep=1pt,circle,draw,thick] (2) at (0.25,0.5) {\small $7$};
\draw[thick] (2) -- (0) -- (1);
\end{tikzpicture}\quad\begin{tikzpicture}
\node[inner sep=0pt,circle,draw,thick,fill=gray] (0) at (0,0) {\small $15$};
\node[inner sep=1pt,circle,draw,thick] (1) at (-0.25,0.5) {\small $3$};
\node[inner sep=1pt,circle,draw,thick] (2) at (0.25,0.5) {\small $5$};
\draw[thick] (2) -- (0) -- (1);
\end{tikzpicture}\quad\begin{tikzpicture}
\node[inner sep=0pt,circle,draw,thick,fill=gray] (0) at (0,0) {\small $16$};
\node[inner sep=1pt,circle,draw,thick] (1) at (0,0.5) {\small $2$};
\node[inner sep=1pt,circle,draw,thick] (2) at (0,1) {\small $2$};
\node[inner sep=1pt,circle,draw,thick] (3) at (0,1.5) {\small $2$};
\draw[thick] (0) -- (1) -- (2) -- (3);
\end{tikzpicture}}\]
We imagine that, at each generation, children of a node are represented above it as nodes sorted in increasing order from left to right. The \textit{depth} of a child is the number of edges of the unique path joining it with the root along the tree. It is immediate to see that if $n$ is prime, $\tau(n)=\{n:1\}$ and the corresponding tree consists of a dimer. If we focus on the \textit{topology} of such trees, ignoring therefore the labels appearing in the nodes, the construction above associates to each natural number $n>1$ a rooted planar tree. As the emerging structure is undecorated, different numbers are mapped in the same tree topology, as it appears in the sequence below, where $6$, $10$, $14$ and $15$ are all mapped in the same ``undecorated'' rooted graph:
\[\resizebox{\linewidth}{!}{\begin{tikzpicture}
\node[inner sep=2pt,circle,draw,thick,fill=gray,label=-90:{\small $1$}] (0) at (0,0) {};
\end{tikzpicture}\quad\begin{tikzpicture}
\node[inner sep=2pt,circle,draw,thick,fill=gray,label=-90:{\small $2$}] (0) at (0,0) {};
\node[inner sep=2pt,circle,draw,thick] (1) at (0,0.5) {};
\draw[thick] (0) -- (1);
\end{tikzpicture}\quad\begin{tikzpicture}
\node[inner sep=2pt,circle,draw,thick,fill=gray,label=-90:{\small $3$}] (0) at (0,0) {};
\node[inner sep=2pt,circle,draw,thick] (1) at (0,0.5) {};
\draw[thick] (0) -- (1);
\end{tikzpicture}\quad\begin{tikzpicture}
\node[inner sep=2pt,circle,draw,thick,fill=gray,label=-90:{\small $4$}] (0) at (0,0) {};
\node[inner sep=2pt,circle,draw,thick] (1) at (0,0.5) {};
\node[inner sep=2pt,circle,draw,thick] (2) at (0,1) {};
\draw[thick] (0) -- (1) -- (2);
\end{tikzpicture}\quad\begin{tikzpicture}
\node[inner sep=2pt,circle,draw,thick,fill=gray,label=-90:{\small $5$}] (0) at (0,0) {};
\node[inner sep=2pt,circle,draw,thick] (1) at (0,0.5) {};
\draw[thick] (0) -- (1);
\end{tikzpicture}\quad\begin{tikzpicture}
\node[inner sep=2pt,circle,draw,thick,fill=gray,label=-90:{\small $6$}] (0) at (0,0) {};
\node[inner sep=2pt,circle,draw,thick] (1) at (-0.25,0.5) {};
\node[inner sep=2pt,circle,draw,thick] (2) at (0.25,0.5) {};
\draw[thick] (2) -- (0) -- (1);
\end{tikzpicture}\quad\begin{tikzpicture}
\node[inner sep=2pt,circle,draw,thick,fill=gray,label=-90:{\small $7$}] (0) at (0,0) {};
\node[inner sep=2pt,circle,draw,thick] (1) at (0,0.5) {};
\draw[thick] (0) -- (1);
\end{tikzpicture}\quad\begin{tikzpicture}
\node[inner sep=2pt,circle,draw,thick,fill=gray,label=-90:{\small $8$}] (0) at (0,0) {};
\node[inner sep=2pt,circle,draw,thick] (1) at (0,0.5) {};
\node[inner sep=2pt,circle,draw,thick] (2) at (0,1) {};
\draw[thick] (0) -- (1) -- (2);
\end{tikzpicture}\quad\begin{tikzpicture}
\node[inner sep=2pt,circle,draw,thick,fill=gray,label=-90:{\small $9$}] (0) at (0,0) {};
\node[inner sep=2pt,circle,draw,thick] (1) at (0,0.5) {};
\node[inner sep=2pt,circle,draw,thick] (2) at (0,1) {};
\draw[thick] (0) -- (1) -- (2);
\end{tikzpicture}\quad\begin{tikzpicture}
\node[inner sep=2pt,circle,draw,thick,fill=gray,label=-90:{\small $10$}] (0) at (0,0) {};
\node[inner sep=2pt,circle,draw,thick] (1) at (-0.25,0.5) {};
\node[inner sep=2pt,circle,draw,thick] (2) at (0.25,0.5) {};
\draw[thick] (2) -- (0) -- (1);
\end{tikzpicture}\quad\begin{tikzpicture}
\node[inner sep=2pt,circle,draw,thick,fill=gray,label=-90:{\small $11$}] (0) at (0,0) {};
\node[inner sep=2pt,circle,draw,thick] (1) at (0,0.5) {};
\draw[thick] (0) -- (1);
\end{tikzpicture}\quad\begin{tikzpicture}
\node[inner sep=2pt,circle,draw,thick,fill=gray,label=-90:{\small $12$}] (0) at (0,0) {};
\node[inner sep=2pt,circle,draw,thick] (1) at (-0.25,0.5) {};
\node[inner sep=2pt,circle,draw,thick] (11) at (-0.25,1) {};
\node[inner sep=2pt,circle,draw,thick] (2) at (0.25,0.5) {};
\draw[thick] (2) -- (0) -- (1) -- (11);
\end{tikzpicture}\quad\begin{tikzpicture}
\node[inner sep=2pt,circle,draw,thick,fill=gray,label=-90:{\small $13$}] (0) at (0,0) {};
\node[inner sep=2pt,circle,draw,thick] (1) at (0,0.5) {};
\draw[thick] (0) -- (1);
\end{tikzpicture}\quad\begin{tikzpicture}
\node[inner sep=2pt,circle,draw,thick,fill=gray,label=-90:{\small $14$}] (0) at (0,0) {};
\node[inner sep=2pt,circle,draw,thick] (1) at (-0.25,0.5) {};
\node[inner sep=2pt,circle,draw,thick] (2) at (0.25,0.5) {};
\draw[thick] (2) -- (0) -- (1);
\end{tikzpicture}\quad\begin{tikzpicture}
\node[inner sep=2pt,circle,draw,thick,fill=gray,label=-90:{\small $15$}] (0) at (0,0) {};
\node[inner sep=2pt,circle,draw,thick] (1) at (-0.25,0.5) {};
\node[inner sep=2pt,circle,draw,thick] (2) at (0.25,0.5) {};
\draw[thick] (2) -- (0) -- (1);
\end{tikzpicture}\quad\begin{tikzpicture}
\node[inner sep=2pt,circle,draw,thick,fill=gray,label=-90:{\small $16$}] (0) at (0,0) {};
\node[inner sep=2pt,circle,draw,thick] (1) at (0,0.5) {};
\node[inner sep=2pt,circle,draw,thick] (2) at (0,1) {};
\node[inner sep=2pt,circle,draw,thick] (3) at (0,1.5) {};
\draw[thick] (0) -- (1) -- (2) -- (3);
\end{tikzpicture}}\]
Some properties of the so-constructed sequence of trees have been investigated in the literature, for example with respect to their depth \cite{DeKoninck2024}. On the other hand, each planar tree can be represented as a Dyck path, or equivalently a binary string \cite{Stanley2011}. Such a string, that we will denote by $w(n)$ with respect to the natural number $n$, can be imagined as constructed starting from the root and moving on the exterior of the tree from its leftmost edge: each time an edge is traversed upward we insert a \texttt{1}, each time an edge is traveled downwards we insert a \texttt{0}. In the example above, $n=2^{2^{2\cdot 3}}\cdot 3$ is therefore mapped into the string $w(n)=\mathtt{1110100010}$. As the tree exploration ends when we get back to the root, each of these binary representations necessarily contains as many \texttt{1}s as \texttt{0}s, always starting with \texttt{1} and ending with \texttt{0}, in a way that on the left of each digit in the string there are no more \texttt{0}s than \texttt{1}s, and rooted ordered trees with $m$ edges are in bijection with Dyck words of semilength $m$. In this representation, all prime numbers correspond to \texttt{10}, as they are associated to the tree $\begin{tikzpicture}[baseline={([yshift=-.5ex]current bounding box.center)}]
\node[inner sep=1.5pt,circle,draw,thick,fill=gray] (0) at (0,0) {};
\node[inner sep=1.5pt,circle,draw,thick] (1) at (0,0.25) {};
\draw[thick] (0) -- (1);
\end{tikzpicture}$. Note that such a representation is obviously not in bijection with $\mathds{N}$, nevertheless it contains information on the factorization properties of natural numbers (so that, for example \texttt{10} indicates that a number is prime, or a sequence \texttt{1010\dots10} of $k$ repetitions of \texttt{10} expresses the fact that the number is a \textit{square-free number}, namely a product of $k$ distinct primes): as a result, the sequence of undecorated trees, that we denote $\NT$, can be seen as a (lossy) translation of $\mathds N$ in a ``language'' of Dyck words following its own ``grammatical rules'' \cite{Childress2021}. 
{In fact}, in a series of works \cite{conti2023,Conti2024,CCI2}, it was observed that the sequence $\NT$ exhibits a number of structural properties. 

Each Dyck word appears in the sequence, and it does so infinitely many times. Both properties are a direct consequence of the infinitude of prime numbers, {and imply that the set $\mathds N\mathcal D$ of distinct Dyck words appearing in the sequence, i.e., the \textit{dictionary}, is countably infinite}. Certain finite sequences of consecutive Dyck words occur only once, while
others recur extremely often, and in several cases the distinction can be
settled by elementary arguments. The sequence of Dyck words \texttt{10 10}, for example, requires two
consecutive primes: one of the two is even, so it occurs exactly once, at
$n=2$. The sequence of Dyck words \texttt{1100 10} requires $n=p^q$ with $p,q$ prime and
$n+1$ prime; parity forces $p=2$, and if $q$ were odd then $3$ would divide
$2^q+1>3$, so $q=2$ and the sequence of Dyck words occurs exactly once, at $n=4$. In the
opposite direction, the infinite occurrence of \texttt{10 1100} is
equivalent to the still-open Mersenne conjecture, while the infinitude of Germain primes would imply that \texttt{1010 10} occurs infinitely
often: for such a prime $p>2$ the integer $n=2p$ is a product of two
distinct primes and $n+1=2p+1$ is prime. Finally, \texttt{1100 1100}
requires two consecutive perfect powers and therefore occurs only once, at
$n=8$, a fact established in 2004 with the proof of Catalan's conjecture by
\textcite{Mihailescu2004}.

The sequence $\NT$ is moreover oriented: if one takes a valid sequence of Dyck words and reverses the
order of its words, the resulting string is often invalid, i.e., it never
occurs. For example, \texttt{111000 10} occurs at least twice, at $n=16$ and
$n=256$, corresponding to the Fermat primes $F_2=17$ and $F_3=257$, whereas
the reversed sequence \texttt{10 111000} never appears in $\NT$: it would
require a prime $n$ with $n+1=p^{q^s}$ for primes $p,q,s$, and parity forces
$p=2$, so that $n=2^{q^s}-1$; this is impossible since $2^{q^{s-1}}-1$ is a
proper nontrivial divisor of $2^{q^s}-1$. Whether \texttt{111000 10} occurs
a third time is equivalent to the existence of a Fermat prime $F_r=2^{2^r}+1$
with prime index $r$ beyond $r=2,3$, an open problem in number theory. Further examples of impossible configurations of words include four consecutive square-free numbers, which is ruled out by the simple observation that one of any four consecutive integers is always divisible by 4, and hence not square-free.
\begin{figure}\centering
\includegraphics[width=\linewidth]{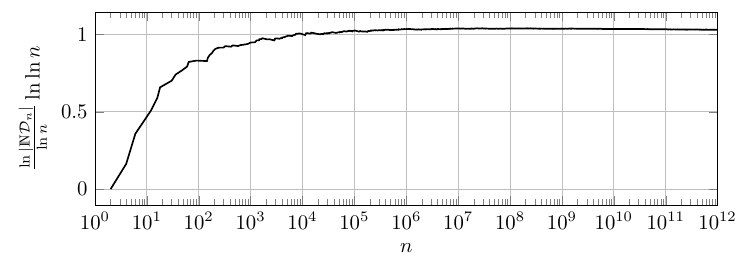}
\caption{Number of elements in the natural dictionary $\ND_n$ as a function of $n$. It is numerically found that $|\ND_n|$ scales sublinearly in $n$ up to $n=10^{12}$ \cite{Modena2025}.}\label{fig:DN}
\end{figure}

Our dataset consists of such a string representation for the first $10^{11}$ natural numbers: as detailed in Section \ref{sec:experiments}, the obtained structured sequence of strings has been fed to a large language model as though it were a long text, with the goal in mind of testing its predictive ability after learning. In the following, we will denote by $\NT_n$ the ordered sequence of undecorated trees corresponding to the first $n-1$ natural numbers larger than $1$ (so that $|\NT_n|=n-1$). Similarly, we denote by $\ND_n$ the set of distinct trees appearing in $\NT_n$, namely the dictionary required to write down $\NT_n$. 

\begin{figure}
    \centering
    \includegraphics[height=0.45\linewidth]{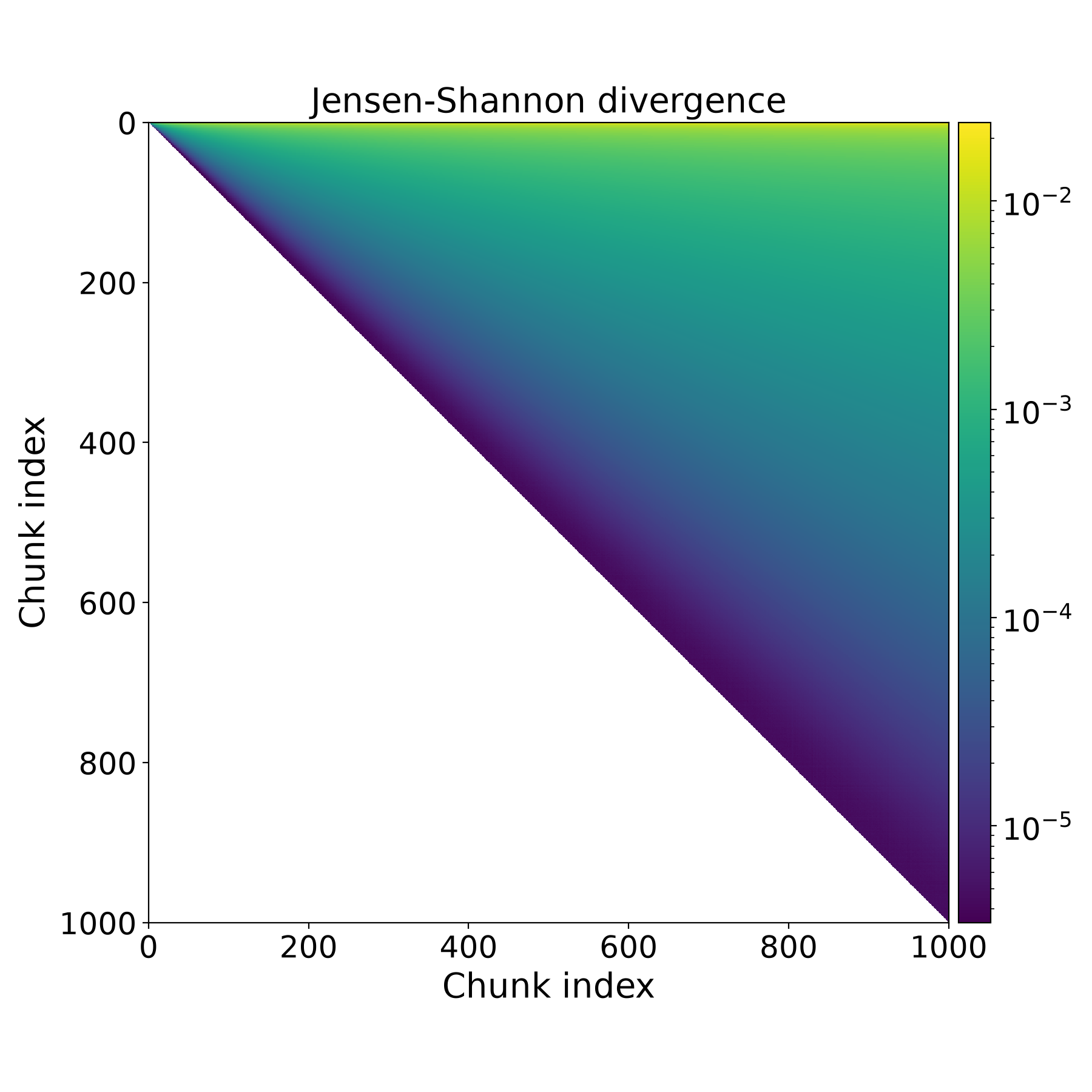}\qquad
    \includegraphics[height=0.45\linewidth]{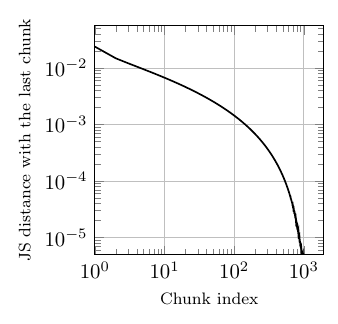}
    \caption{(\textit{Left}) Jensen--Shannon divergence between chunks of $10^8$ consecutive Dyck words in our dataset $\NT_n$ with $n=10^{11}$. The largest value of the divergence, reached between the very first chunk and the very last (upper right corner), is approximately equal to $0.0238$, i.e., the distribution difference remains quite small despite the $11$ orders of magnitude span. (\textit{Right}) Detail of the Jensen--Shannon divergence of the last chunk of $10^8$ trees with respect to all the previous ones of equal size.}
    \label{fig:jensen}
\end{figure}

The statistical features of the database have been analyzed 
in \cite{Modena2025}. That study provides some evidence that the arithmetic structure underlying $\NT$ produces an organized and self-consistent symbolic sequence. The number $|\ND_n|$ of distinct symbolic units grows sublinearly with $n$, indicating the emergence of internal rules and recursive patterns, see Fig.~\ref{fig:DN}. The statistical properties of $\NT$ are \textit{not} invariant under translations along the sequence, and the distribution over the possible trees is subject to a deformation with the scale. In Fig.~\ref{fig:jensen} we exemplify this fact: we divided the sequence $\NT_n$ with $n=10^{11}$ into $10^3$ chunks of $10^{8}$ consecutive trees and we compared, using the Jensen--Shannon divergence, the associated frequency distributions over $\ND_n$. We recall that, given two generic distributions $p$ and $q$ on a finite set $\Omega$, the Jensen--Shannon divergence $\mathrm{JS}(p\|q)$ quantifies their similarity and it is defined as $$\mathrm{JS}(p\|q)\coloneqq\frac{1}{2}\mathrm{KL}_2(p\|m)+\frac{1}{2}\mathrm{KL}_2(q\|m)\in[0,1],$$ where $\mathrm{KL}_2(p\|q)\coloneqq\sum_{x\in\Omega} p(x)\log_2\frac{p(x)}{q(x)}$ is the Kullback–Leibler divergence with base-2 logarithm and $m(x)\coloneqq\frac{p(x)+q(x)}{2}$ is a mixture distribution on the same set $\Omega$. It is visible that this quantity is clearly nonzero and larger for distant chunks, although it takes values below $0.025$ for all pairs (with the maximum divergence occurring between the first and the last chunks). Although clearly statistically non-homogeneous, $\NT$ shows a degree of approximate homogeneity within a good number of orders of magnitude.

Finally, observe that the sequence is inherently oriented, with directionality embedded in its ``syntax'', and shows a balance between redundancy and novelty as revealed by entropy and compression analyses \cite{Modena2025}. The rank–frequency hierarchy remains nearly invariant across scales, while correlation studies expose long-range dependencies and a transition from diffusive to ballistic regimes. These similarities to natural languages naturally lead to the question of whether, how, and to what extent the structure of $\NT$ can be learned by a large language model, the central question addressed in this paper.

\section{The Learning Model}\label{sec:model}

{Transformers}~\cite{vaswani2017attention} are a
powerful type of neural network that is currently achieving state-of-the-art results in various domains, such as natural language processing (NLP)~\cite{devlin2018bert, howard2018universal,
  radford2018improving, brown2020language, chatgpt},  image
classification~\cite{dosovitskiy2021an}, protein structure
prediction~\cite{jumper2021highly}, detection of ground states of many-body quantum
systems~\cite{viteritti2022transformer, rende2023simple, viteritti2023transformer} and many others. Contrary to previous models of neural networks, transformers have been shown to be better suited to capture long-range correlations in sequential data, like, for instance, Dyck words in a sequence. This peculiar property of such architectures motivated us to test their effectiveness in capturing correlations and structure in a special, structured sequence of symbols determined by an underlying complex set of rules, such as $\mathds N\mathcal T$.   

In light of the subsequent discussion, let us now review some essentials about transformers. In the transformer processing, the machine deals with $D$ possible pre-assigned \textit{token values}, each corresponding to a vector in $d$ dimensions, representing the \textit{dictionary} $\mathcal D$ the transformer uses. The text sequence the transformer has to process (in our case, the sequence of Dyck words) is therefore first mapped, through a \textit{tokenizer}, into an ordered sequence of such $d$-dimensional tokens: it is now well known that this continuous embedding space is essential for enabling generalization across related symbols rather than treating each token as an unrelated categorical label \cite{embedding_reprs}. Given an ordered sequence of $L$ tokens, $\bt=(\bt_1,\dots,\bt_L)$, the self-attention layer transforms each token in the sequence, let us call it $\bt_i\in\R^d$, into a vector $\bh_i\in\R^d$, a combination of information from the other tokens in the sentence. Such linear combination is obtained via an \textit{attention matrix} $\mathbf A \in \R^{L \times L}$, built in such a way that $A_{ij}$ describes how relevant the token $\bt_j$ is to understanding the semantic meaning of the token $\bt_i$: 
\begin{equation}
   \bh_i \coloneqq \sum_{j=1}^L
   A_{ij}\boldsymbol{v}_j\in\R^d.
\end{equation}
The linear projection $\boldsymbol{v}_j\coloneqq\mathbf V \bt_j$ is called \emph{value} and is parameterized by a matrix $\mathbf V \in \R^{d \times d}$ that is learned during training. The attention weights, on the other hand, are computed by estimating a similarity score between the $i$-th token and all the other tokens in the sequence:
\begin{equation}
    A_{ij} \coloneqq \frac{\exp\left(\bq_i^\intercal\bk_j \right)}{\sum_r\exp\left(\bq_i^\intercal\bk_r\right)}\in[0,1].
\end{equation}
The $i$th row of the attention matrix is therefore obtained via a softmax on the vector $(\bq_i^\intercal\bk_j)_j$. The linear projections $\bq_i \coloneqq \mathbf Q \bt_i$ and $\bk_i \coloneqq \mathbf K \bt_i$ are called \emph{query} and \emph{key}, respectively, and result from the application of two trainable matrices, $\mathbf Q \in \R^{R \times d}$ and $\mathbf K \in \R^{R \times d}$, to the tokens: here $R$ is a hyperparameter to be fixed by fine-tuning. As a result, the vector $\bh_i$ contains information concerning the entire sequence, hopefully weighting each token in it depending on its relevance for determining the token $\bt_i$. A pictorial representation of this mechanism is given in Fig.~\ref{fig:attention}.

\begin{figure}[!tb]

\begin{center}
\newcommand{\alberoA}{%
  \begin{tikzpicture}[baseline=0pt, scale=0.7, every node/.style={transform shape}]
\node[inner sep=2pt,circle,draw,thick,fill=gray] (0) at (0,0) {};
\node[inner sep=2pt,circle,draw,thick] (1) at (0,0.5) {};
\draw[thick] (0) -- (1);
  \end{tikzpicture}%
}
\newcommand{\alberoB}{%
  \begin{tikzpicture}[baseline=0pt, scale=0.7, every node/.style={transform shape}]
  \node[inner sep=2pt,circle,draw,thick,fill=gray] (0) at (0,0) {};
\node[inner sep=2pt,circle,draw,thick] (1) at (-0.25,0.5) {};
\node[inner sep=2pt,circle,draw,thick] (2) at (0.25,0.5) {};
\draw[thick] (2) -- (0) -- (1);
  \end{tikzpicture}%
}  
\newcommand{\alberoC}{%
  \begin{tikzpicture}[baseline=0pt, scale=0.7, every node/.style={transform shape}]
    \node[inner sep=2pt,circle,draw,thick,fill=gray] (0) at (0,0) {};
    \node[inner sep=2pt,circle,draw,thick] (1) at (-0.25,0.5) {};
    \node[inner sep=2pt,circle,draw,thick] (11) at (-0.25,1) {};
    \node[inner sep=2pt,circle,draw,thick] (2) at (0.25,0.5) {};
    \draw[thick] (2) -- (0) -- (1) -- (11);
  \end{tikzpicture}%
}
\newcommand{\alberoD}{%
  \begin{tikzpicture}[baseline=0pt, scale=0.7, every node/.style={transform shape}]
\node[inner sep=2pt,circle,draw,thick,fill=gray] (0) at (0,0) {};
\node[inner sep=2pt,circle,draw,thick] (1) at (0,0.5) {};
\node[inner sep=2pt,circle,draw,thick] (2) at (0,1) {};
\draw[thick] (0) -- (1) -- (2);
  \end{tikzpicture}%
}
\resizebox{\linewidth}{!}{\begin{tikzpicture}[remember picture,
    outerbox/.style={
        draw=black, very thick,
        rectangle, 
        inner sep=1pt,
        node distance=1pt
    },
    innerbox/.style={
        fill=gray!20,
        draw=gray!50,
        rectangle,
        inner sep=2pt,
        font=\ttfamily
    },
    start chain=going right
]

\node[outerbox, on chain] (n1) {\tikz\node[innerbox] (V5) {10};\tikz\node[innerbox]{\phantom{0} };\tikz\node[innerbox] (V6){1010};\tikz\node[innerbox]{\phantom{0} };};

\node[outerbox, on chain] (n2) {\tikz\node[innerbox] (V7) {10};\tikz\node[innerbox]{\phantom{0} };\tikz\node[innerbox] (V8){1100};};
    
\node[outerbox, on chain] (n3) {\tikz\node[innerbox]{\phantom{0} };\tikz\node[innerbox] (V9){1100};};

\node[outerbox, on chain] (n4) {\tikz\node[innerbox]{\phantom{0} };\tikz\node[innerbox] (V10){1010};};
    \node[outerbox, on chain] (n5) {\tikz\node[innerbox]{\phantom{0} };}; 
    \node[outerbox, on chain] (n6) {\tikz\node[innerbox] (V11){10};\tikz\node[innerbox]{\phantom{0} };\tikz\node[innerbox] (V12){1100};};
    \node[outerbox, on chain] (n7) {\tikz\node[innerbox]{10};\tikz\node[innerbox]{\phantom{0} };};
    \node[outerbox, on chain] (n8) {\tikz\node[innerbox] (V13){10};\tikz\node[innerbox]{\phantom{0} };\tikz\node[innerbox] (V14){1010};};
    
\foreach \x in {1,..., 8} {\node[anchor=north west,yshift=2pt] at (n\x.south west) {$\bt_\x$};}
\draw[stealth-,line width=0.1pt] (n1.south) to [out=-90,in=-90] node[fill=white] {$A_{81}$} (n8.south);
\draw[stealth-,line width=0.1pt] (n2.south) to [out=-90,in=-90] node[fill=white] {$A_{82}$}(n8.south);
\draw[stealth-,line width=1.25pt] (n3.south) to [out=-90,in=-90] node[fill=white] {$A_{83}$}(n8.south);
\draw[stealth-,line width=0.5pt] (n4.south) to [out=-90,in=-90] node[fill=white] {$\dots$}(n8.south);
\draw[stealth-,line width=0.2pt] (n5.south) to [out=-90,in=-90] (n8.south);
\draw[stealth-,line width=0.4pt] (n6.south) to [out=-90,in=-90] (n8.south);
\draw[stealth-,line width=2pt] (n7.south) to [out=-90,in=-90] (n8.south);
\foreach \x in {5,7,11,13} {
\node[anchor=south, align=center, inner sep=0pt] at (V\x.north) {\alberoA\\[-2pt]{\color{gray} $\x$}\\\vspace{-0.1cm}};
}
\foreach \x in {6,10,14} {
\node[anchor=south, align=center, inner sep=0pt] at (V\x.north) {\alberoB\\[-2pt]{\color{gray} $\x$}\\\vspace{-0.1cm}};
}

\foreach \x in {8,9} {
\node[anchor=south, align=center, inner sep=0pt] at (V\x.north) {\alberoD\\[-2pt]{\color{gray} $\x$}\\\vspace{-0.1cm}};
}
\node[anchor=south, align=center, inner sep=0pt] at (V12.north east) {\alberoC\\[-2pt]{\color{gray} $12$}\\\vspace{-0.1cm}};
\end{tikzpicture}}
\end{center}
\caption{An example of a numerical sequence (integers between $5$ and $14$) converted into a sequence of Dyck words. The tokenizer splits the sequence of $10$ Dyck words into $8$ tokens, corresponding to the black boxes. Note that a token can consist of a single Dyck word, a concatenation of Dyck words or, generically, parts of Dyck words (in the example, the Dyck word \texttt{110010} representing $12$ is split into two different tokens). After training, the attention matrix $\mathbf A$ can be used to compute, within a sequence, the relevance of each element of the context for a given token. In the picture, we focus on the last token, containing the Dyck words representing $13$ and $14$. Arrows point towards the other tokens, with width proportional to the associated attention weight: thicker lines point towards more relevant tokens in determining its value. In the example, tokens $\bt_3$ and $\bt_7$ are the most relevant for $\bt_8$, as $A_{83}$ and $A_{87}$ are the largest entries of the attention matrix.}
    \label{fig:attention}
\end{figure}

Modern architectures use multiple \textit{attention heads}. Multi-head attention extends the aforementioned computations by applying multiple attention operations in parallel of self-attention layers, in which each head should ideally capture different patterns and relationships between tokens, allowing for more complex representations. A complete transformer block is thus composed of a multi-head self-attention layer, a normalization regularizer layer and a feed-forward layer~\cite{vaswani2017attention}. Current models stack sequentially several of these blocks to create more refined and rich representations. The attention output $\bh_i$ is transformed, via these additional layers, into an ``unembedded'' vector $\hat\bh_i\in\R^D$ with the dictionary dimensionality that is informative about the statistical properties of the token $\bt_i$ given the context provided by the sentence $\bt$. In particular, $\hat\bh_i$ parametrizes the probability of observing a given element of the dictionary in position $i$ given the surrounding context \cite{rende2024distributional,rende2024mapping}. Details on the training process are given in \ref{app:training}.

The model used in the present work is a \textit{transformer decoder}, inspired by the OpenAI GPT-2 architecture \cite{radford2019language}. The largest model we implemented is an architecture with 12 layers, each combining sequentially multi-head attention, layer normalization and finally a feed-forward neural network adopting GeLU activation function. A single head has dimension 64, and since the multi-head attention is composed of 12 heads, this leads to an embedding dimension $d=768$ adopted in this manuscript. The resulting architecture has a total of $P=8.7\cdot 10^7$ trainable parameters. In the following, we will generically denote by $\btheta$ the set of parameters. We refer to the cited literature for the additional technical details about this specific architecture.

Fixing this large number of parameters requires training via self-supervised tasks, that is, tasks where the label is a part of the input itself. Some examples are \emph{Masked Language Modelling} (MLM) or \emph{Next-Token Prediction} (NTP). In the first case, the Transformer is trained to identify missing tokens within a text. In the second, it is tasked with predicting the next token in a sentence, given its preceding context. In the case of causal-attention for NTP, implemented to model causally sequential data (as text), the upper triangular part of the attention matrix is set to $-\infty$ before the softmax step. In such a way, tokens can only look for correlations with previous ones. By doing so, the model selectively aggregates information based on context, learning representations that are independent of tokens positions. Therefore, positional embeddings are added to the token embeddings to characterize the positional information of data, that is not captured by attention on its own. 

MLM and NTP are precisely the two tasks on which we will test the ability of a Transformer to learn the internal structure of $\NT$, as we will clarify in the next Section.

\section{The Explored Tasks}\label{sec:tasks}
We trained our LLM architecture by focusing on two specific tasks. In both cases, the main goal is to evaluate whether and how much the architecture is able to learn the ``syntactical rules'' underlying the sequence $\NT$. As anticipated, our base dataset consists of the ordered sequence of the first $n$ trees $\NT_n$, represented via Dyck words. This dataset is preprocessed and converted into a different dataset ${\mathcal T}_n$ via a \textit{tokenizer} (see Section \ref{sec:token}): our architecture will be fed the sequences in the tokenized representation. In particular, we construct a set ${\mathcal T}_{\rm train}$ of sequences of $L$ consecutive tokens to be used for training, in the form $\bt=(\bt_i)_{i=1}^L\in{\mathcal T}_{\rm train}$, corresponding, via the tokenizer, to a sequence of consecutive number representations in $\NT$. In what follows, a \textit{tokenized sentence} denotes a contiguous sequence of $L$ BPE tokens used as a training or evaluation example. Such a sentence does not necessarily contain $L$ complete Dyck words, because token boundaries and Dyck-word boundaries need not coincide. The term \textit{prompt} is reserved for a token sequence supplied to the model as input during inference or likelihood evaluation. Similarly, we constructed a set ${\mathcal T}_{\rm test}$ of sequences of $L$ consecutive tokens to be used for testing, in such a way that ${\mathcal T}_{\rm test}\cap{\mathcal T}_{\rm train}=\varnothing$. The precise characterization of the training and test datasets will be given in Section~\ref{sec:experiments}. The goal of the training phase is then to find a configuration of the attention weights and values that achieve optimal generalization performance on previously unseen tokenized sentences during the training phase, namely sequences in ${\mathcal T}_{\rm test}$. The training process, and the following evaluation, depends on the \textit{specific task} the model is trained to perform.

\paragraph{Next-Token Prediction} Let us consider the sentence $\bt=(\bt_i)_{i=1}^L\in{\mathcal T}_{\rm train}$ in the training dataset, and let $\bt_{1:k}\coloneqq(\bt_i)_{i=1}^{k}$ be the sub-sentence consisting of the first $k$ tokens. In a Next-Token Prediction (NTP) task we train our architecture via gradient descent to predict the next token $\bt_{k+1}$ of the sequence. To do so, we aim at minimizing a cross-entropy loss, i.e.,
\begin{equation}
    \mathcal{L}_{\rm n}(\boldsymbol\theta) \coloneqq - \frac{1}{|{\mathcal T}_{\rm train}|}\sum_{\bt\in{\mathcal T}_{\rm train}}\frac{1}{L-1} \sum_{k=1}^{L-1}
 \ln p_{\boldsymbol\theta}(\bt_{k+1}|\bt_{1:k})\label{eq:nwp_loss}
\end{equation}
in which $p_{\boldsymbol\theta}(\bt_{k+1} |\bt_{1:k}) \in [0,1] $ is the transformer's output probability for the correct next token $\bt_{k+1}$ given the previous $k$ tokens in the sequence $\bt$, that is, given the vector $\bt_{1:k}$. Such a probability depends on the parameters $\btheta$ required by the architecture and is obtained by applying the softmax activation function to the pre-activation of the last layer $\hat\bh_{k+1}\equiv\hat\bh_{k+1}(\bt_{1:k})\in\R^D$ depending on the input sequence $\bt_{1:k}$, so that the predicted probability distribution for the value of $\bt_{k+1}$ is
\begin{equation}
p_{\boldsymbol{\theta}}(\bt_{k+1}=\bt_a|\bt_{1:k}) \coloneqq\frac{\exp(\beta \hat h_{k+1,a})}{\sum_{b=1}^{D}\exp(\beta \hat h_{k+1,b})},\qquad a\in[D].\label{eq:soft}
\end{equation}
In the expression above, an \textit{inverse temperature} parameter $\beta$ appears: as it will be clear below, its value is very relevant in the final performance of the LLM. 

\paragraph{Masked Language Modeling}  Let $\bt=(\bt_i)_{i=1}^L$ be a sequence of $L$ consecutive tokens, and let $\{\bt_a\}_{a\in\mathcal I(\bt)}$ be a subset of tokens appearing in $\bt$, so that $\mathcal I(\bt)\subset[L]$ is an index set consisting of non-repeated randomly selected indices, which we assume being $15\%$ of the original sequence. The set $\mathcal I(\bt)$ will identify the subset of ``masked tokens''. In the Masked Language Modeling task, each masked element $\bt_a$, $a\in\mathcal I(\bt)$, is randomly replaced in the sequence by a special mask-element token \texttt{M} (with probability $0.75$), by a different token randomly drawn from the dictionary (with probability $0.15$), and is left unchanged otherwise. The masking probability value is a standard protocol in the literature \cite{devlin2018bert}. We train a transformer via gradient descent to predict the possibly missing token $\bt_a$, $a\in\mathcal I(\bt)$, given the altered sequence $\bt_{\not a}$ obtained from $\bt$ by the described masking procedure on the token $\bt_a$ only. The following cross-entropy loss
\begin{equation}\label{eq:trainloss}
    \mathcal{L}_{\rm m}(\boldsymbol\theta) \coloneqq - \frac{1}{|{\mathcal T}_{\rm train}|}\sum_{\bt\in{\mathcal T}_{\rm train}}\frac{1}{|\mathcal I(\bt)|} \sum_{a\in\mathcal I(\bt)}
 \ln p_{\boldsymbol\theta}(\bt_a |\bt_{\not a})
\end{equation}
is used for training, where the role of ``true label'' is taken, for each $a$, by the true missing token $\bt_a$ itself, while $p_{\boldsymbol\theta}(\bt_a|\bt_{\not a}) \in [0,1] $ represents the transformer's output probability about the true, missing token given all the others in the sentence that represent the context. This quantity depends of course on the parameters in the architecture. Each output probability $p_{\boldsymbol\theta}(\cdot|\bt_{\not a})$ is obtained by applying the softmax activation function to the pre-activation of the last layer $\hat\bh\coloneqq \hat\bh(\bt_{\not a})\in\R^D$ depending on the input masked sentence $\bt_{\not a}$ in the very same form as in Eq.~\eqref{eq:soft}. Note that no causal masking is applied to the attention matrix during the MLM training.

\section{Results of the Experiments}
\label{sec:experiments}
We conducted an extensive experimental evaluation to test the performance of the model described in Section~\ref{sec:model} on two distinct training strategies, namely the Next-Token prediction task and the masked language modeling. To this end, we employed the large (ordered) dataset $\NT_n$ described in Section~\ref{sec:data} that consists of the sequence of Dyck words that represent all the integer numbers from $2$ up to $n=10^{11}$. Such a dataset has been split into training and test sets, pre-processed via a tokenization process to produce a new representation in terms of \textit{tokens}. Before presenting the performed experiments, it is therefore convenient to describe how we managed the dataset, and how we performed such a tokenization.

\paragraph{Dataset Management} The ordered dataset $\NT_n$, with $n=10^{11}$, was first split into $10$ chunks of $10^{10}$ Dyck words each (except for the first, consisting of $10^{10}-1$ entries).
\begin{figure}
\begin{tikzpicture}[scale=0.9]
\foreach \x in {1,...,10} {
  \draw[gray] ({(\x*1.1-1.1-0.025)}, -0.03) rectangle ++(1.04, 0.34);
  \node[draw, fill=blue!10, minimum width=0.75cm, minimum height=0.3cm, anchor=south west] (B\x) at ({(\x*1.1-1.1)}, 0) {};
  \node[draw, fill=red!10, minimum width=0.25cm, minimum height=0.3cm, anchor=south west] (V\x) at ({(\x*1.1-1.1+0.75)}, 0) {};
  
  \pgfmathparse{\x==1 ? "10^{10}-1" : "10^{10}"}%
  \node[anchor=south] at (B\x.north) {\tiny\color{gray} $\pgfmathresult$};
}
\node[draw, fill=red!30, minimum width=0.25cm, minimum height=0.3cm, anchor=south west] (Vbb) at (10*1.1-1.1+ 0.75, 0) {};
\node[draw, fill=red!30, minimum width=0.1cm, minimum height=0.3cm, anchor=south west] (Vb1) at (10*1.1-1.1+ 0.75+0.5, 0) {};
\node[draw, fill=red!30, minimum width=0.1cm, minimum height=0.3cm, anchor=south west] (Vb2) at (10*1.1-1.1+ 0.75+1, 0) {};
\node[draw, fill=red!30, minimum width=0.1cm, minimum height=0.3cm, anchor=south west] (Vb3) at (10*1.1-1.1+ 0.75+1.5, 0) {};
\node[anchor=west, rotate=-90] at (Vb1.south) {\tiny\color{gray} $10^{6}$ at $10^{13}$};
\node[anchor=west, rotate=-90] at (Vb2.south) {\tiny\color{gray} $10^{6}$ at $10^{14}$};
\node[anchor=west, rotate=-90] at (Vb3.south) {\tiny\color{gray} $10^{6}$ at $10^{15}$};

\node (A) at ($(B5) + (0, -1)$) {\small Training};
\node (N) at ($(B1) + (-1.3, 0)$) {\small $\mathds N\mathcal T_n=$};
\node (Vn) at ($(V5) + (0, 1)$) {\small Validation};
\node (VT) at ($(V10) + (0, 1)$) {\small Test};
\foreach \x in {1, ..., 4} {
\draw[looseness=0.75,-stealth] (B\x.south) to[out=-90,in=160] (A);
\draw[looseness=0.75,-stealth] (V\x.north) to[out=90,in=200] (Vn);
    }
\draw[-stealth] (B5.south) to (A.north);
\draw[-stealth] (V5.north) to (Vn.south);
\draw[-stealth] (V10.north) to (VT.south);
\draw[-stealth] (Vb1.north) to[out=90,in=-90] (VT.south);
\draw[-stealth] (Vb2.north) to[out=90,in=-90] (VT.south);
\draw[-stealth] (Vb3.north) to[out=90,in=-90] (VT.south);
\foreach \x in {6, ..., 10} {
\draw[looseness=0.75,-stealth] (B\x.south) to[out=-90,in=20] (A);}
\foreach \x in {6, ..., 9} {
\draw[looseness=0.75,-stealth] (V\x.north) to[out=90,in=-20] (Vn);
    }
\end{tikzpicture}  
\caption{Scheme of the dataset management. The sequence of $10^{11}-1$ consecutive Dyck words has been split in $10$ blocks of $10^{10}$ Dyck words (except for the first block, that is long $10^{10}-1$), each further separated into two portions: the first $\sfrac{3}{4}$ of each block is used for training and the rest for validation, except for the last sub-block used for testing. Three additional blocks of $10^6$ consecutive elements were produced for testing in the range $[10^{13},10^{13}+10^6]$, $[10^{14},10^{14}+10^6]$ and $[10^{15},10^{15}+10^6]$ respectively.}
\label{fig:dataset}
\end{figure}
Each one of the first $9$ chunks was further split into the initial $\sfrac{3}{4}$ for parameters training, and the remaining $\sfrac{1}{4}$ for validation (i.e., for the determination of the hyper-parameters of the model). We adopted this splitting of the dataset so that the validation set is more consistent with the training set in terms of statistical properties, thus representing a more robust evaluation set for the hyper-parameter selection phase. The first $\sfrac{3}{4}$ of the $10$th chunk is also used to complement the training set $\mathcal T_{\rm train}$. The remaining $\sfrac{1}{4}$ of this chunk, instead, has been used for testing the performance of the architecture and used to construct $\mathcal T_{\rm test}$. In addition to this block, three additional sequences of $10^6$ consecutive trees were produced for the range $[10^k,10^k+10^6]$ with $k=13,14,15$. The sentences obtained from these batches, far beyond the training set scale, will be added to the test set $\mathcal T_{\rm test}$ as well. The constructed test set can have therefore slightly different statistical properties with respect to the previous training sequences, and it offers a very interesting benchmark to verify the accuracy of the model in predicting unseen intervals at arbitrary distance from our training dataset. A pictorial representation of this dataset use can be found in Fig.~\ref{fig:dataset}. The feeding of the dataset into our architecture, however, is not straightforward, and the construction of the actual sets $\mathcal T_{\rm train}$ and $\mathcal T_{\rm test}$ requires the preprocessing of the Dyck words sequences via a \textit{tokenization} procedure, which we describe in the next paragraph.

\paragraph{Tokenization}\label{sec:token}
As anticipated, to input our dataset in our architecture we need to transform text into a machine-readable format. A widely adopted solution to do so is the so-called \textit{subwords tokenization}. This strategy introduces an intermediate symbolic layer by segmenting text into subword units, enhancing the expressivity of the character-based tokenization and optimizing the large dictionary of word-based encoding. 

Let $(w_1,w_2,\dots,w_n)$ be the sequence of Dyck words in our dataset, of length $n$. This sequence is represented as a sequence of characters from the finite and fixed text alphabet $\mathcal A\coloneqq\{[\mathtt{0}],[\mathtt{1}], [\rule{2\dunder}{0.4pt}] \}$ consisting of \texttt{0}, \texttt{1} and space. A first step in the tokenization is to create an adequate \textit{BPE dictionary} $\mathcal D$ of a given dimension $D\coloneqq|\mathcal D|$. The size $D$ of the tokenizer dictionary is a tunable hyperparameter: in our experiments, for example, we considered $D\in\{2^6,2^8,2^{10}\}$. Note that a ``natural dictionary'' might simply consist of the $|\mathds N\mathcal D_n|$ different Dyck words appearing in our training dataset: for $n=10^{11}$, this would have required $D>2000$, which is a high number but nevertheless manageable. Using as dictionary the ``words'' appearing in our dataset has, however, an important drawback, as it does not allow the architecture to construct unseen words, an undesired feature as we know that infinite Dyck words appear in the sequence $\mathds N\mathcal T$. In our experiments, we have therefore opted for the construction of the dictionary via the Byte-Pair Encoding algorithm \cite{BPE}, which will in fact allow for the creation of unseen sequences. The algorithm constructs $\mathcal D$ iteratively, starting from $\mathcal D=\mathcal A$, set of all characters. At each iteration $k$, the algorithm computes the frequency $f_k(\mathtt{a},\mathtt{b})$ of all adjacent symbol pairs 
$(\mathtt{a},\mathtt{b}) \in \mathcal D \times \mathcal D$ occurring in the corpus. Then, the pair with the highest frequency, $(\mathtt{a}^*,\mathtt{b}^*) =  \arg\max_{(\mathtt{a},\mathtt{b})}f_k(\mathtt{a},\mathtt{b})$, enters as a new, single element of the dictionary $[\mathtt{t}]\coloneqq[\mathtt{a}^*\mathtt{b}^*]$. The vocabulary $\mathcal D$ is updated as $\mathcal D \mapsto \mathcal D \cup \{[\mathtt{t}]\}$. This process is iterated until the desired tokenizer dictionary size $D$ is reached. This tokenization ensures lossless encoding/decoding and high compression thanks to the hierarchical, frequency-based structure.

Once the dictionary $\mathcal D=\{[\mathtt{t}^\nu]\}_{\nu=1}^D$ is constructed, each token in it is associated to a one-hot representation, $[\mathtt{t}^\nu]\mapsto \btau^\nu=(\delta_{ij})_{j=1}^D\in\R^D$ and then, by means of a (trainable) embedding matrix $\mathbf W_e\in\R^{d\times D}$, mapped into a $d$-dimensional space, so that $[\mathtt{t}^\nu]\mapsto \bt^\nu\coloneqq\mathbf W_e\btau^\nu$, vector to be fed to the attention layers.

A sequence of $\kappa$ Dyck words in the dataset is therefore mapped into a new string of length $K\neq \kappa$ of $d$-dimensional vectors corresponding to the elements of the tokenizer dictionary\footnote{In the following, we will not distinguish between an element $[\mathtt{t}_{a}]$ of the dictionary and its representation $\bt_a\in\R^d$.}
$$(w_1,w_2,\dots,w_{\kappa})\mapsto ([\mathtt{t}_{1}],[\mathtt{t}_{2}],\dots [\mathtt{t}_{K}])\mapsto (\bt_1,\bt_2,\dots,\bt_{K}).$$ 
In this way, we processed the training set in our dataset and obtained a sequence of tokens in place of it. This sequence was further split into consecutive tokenized sentences, each containing  $L=2^{10}$ tokens. As anticipated, we called the resulting set of token sequences ${\mathcal T}_{\rm train}$. Similarly, the test set ${\mathcal T}_{\rm test}$ was obtained by tokenizing the last portion of the $10$th chunk of the entire dataset and the additional batches at larger order of magnitudes, as described above. The sentence length $L$ has been kept fixed for all runs of the experiments, unless otherwise specified.

\paragraph{Training} Once the dataset has been tokenized, training by using the suitable loss fixes the parameters and the hyperparameters of the model. The training took place by using $\mathcal T_{\rm train}$ for up to $15,000$ training steps adopting \texttt{AdamW} as optimizer and a cosine learning-rate schedule with linear warm-up. Training is performed at $\beta=1$, following the standard \texttt{PyTorch} implementation~\cite{pytorch}. Candidate learning rates were evaluated over $750$ training steps, and the value yielding the lowest validation loss at the end of training was selected for the subsequent full training run. As customary with neural architectures~\cite{bengio2012practical}, the training phase exploits an early stopping protocol on the loss computed over the validation set, with a \textit{patience value} equal to 6, i.e., training continues until the validation loss sees no improvement for 6 consecutive training steps.

\subsection{A Baseline: The Hidden Markov Model}

As a baseline for our prediction tasks we considered a Hidden Markov Model (HMM) \cite{rabiner1989tutorial}, a minimal probabilistic model for sequence prediction with latent states. Let $D$ be the size of the tokenizer dictionary $\mathcal D=\{[\mathtt{t}_i]\}_{i=1}^D$, as before. The model consists of a transition matrix $\boldsymbol{\Phi}=(\Phi_{ij})_{ij}\in\R^{H\times H}$, where $H$ is the dimension of the latent space, an \textit{emission matrix} $\mathbf{E}=(E_{ij})_{ij}\in\R^{D\times H}$, and an initial hidden state distribution $\boldsymbol p\in\R^H$. Each element $\Phi_{ij}$ expresses the probability of transitioning from the $j$-th hidden state to the $i$-th hidden state, whereas an element $E_{ij}$ expresses the probability of observing $[\mathtt{t}_i]$ in the hidden state $j$. The $H(H-1)+H(D-1)+(H-1)$ free parameters of the model have been determined by using the \texttt{ADAM} optimizer on the empirical average negative log-likelihood loss for the NTP task with $L=1$, computed on the very same training dataset used for the LLM training. The performance of the HMM will be used as minimal baseline for our results, as we expect our LLM to be \textit{at least} as good as a standard hidden Markovian model. For each choice of $D$, we implemented a HMM with $H=2D$, providing sufficient representational capacity to capture multiple latent regimes while avoiding the excessive complexity associated with much larger hidden spaces.

\subsection{Next-Token Prediction}

Let us start by presenting the performance of the model trained on the NTP task. As anticipated, the prompt is constructed with a context of $L=2^{10}$ BPE tokens, with no further instruction. The LLM then generates the continuation of the sequence.
\begin{figure}
    \centering
    \includegraphics[width=\linewidth]{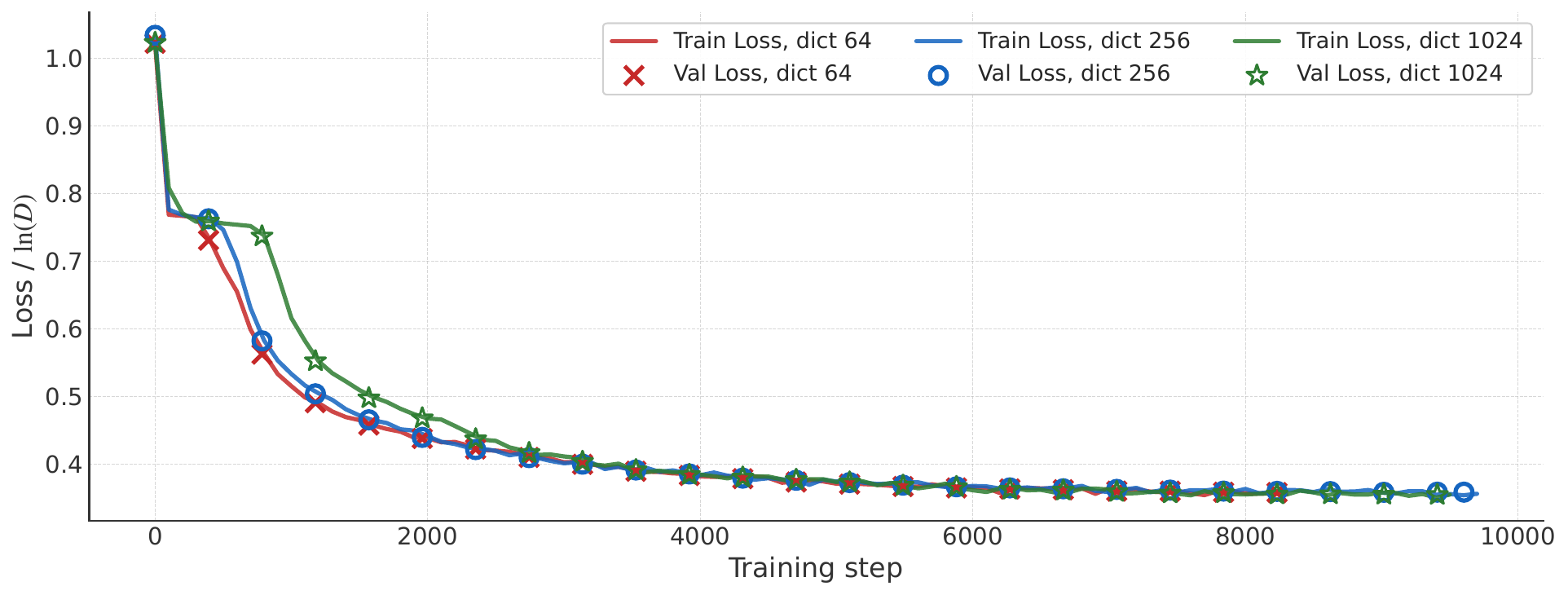}
    \caption{Loss curves for a model trained via NTP, obtained for different tokenizer's vocabulary size $D$, namely $D=2^6$, $D=2^8$, $D=2^{10}$. The loss value, computed as in Eq.~\eqref{eq:nwp_loss}, is rescaled by $\ln D$, loss value corresponding to a uniform distribution.}
    \label{fig:nwp_losses}
\end{figure}

\begin{figure}
    \centering
    \includegraphics[width=0.95\linewidth]{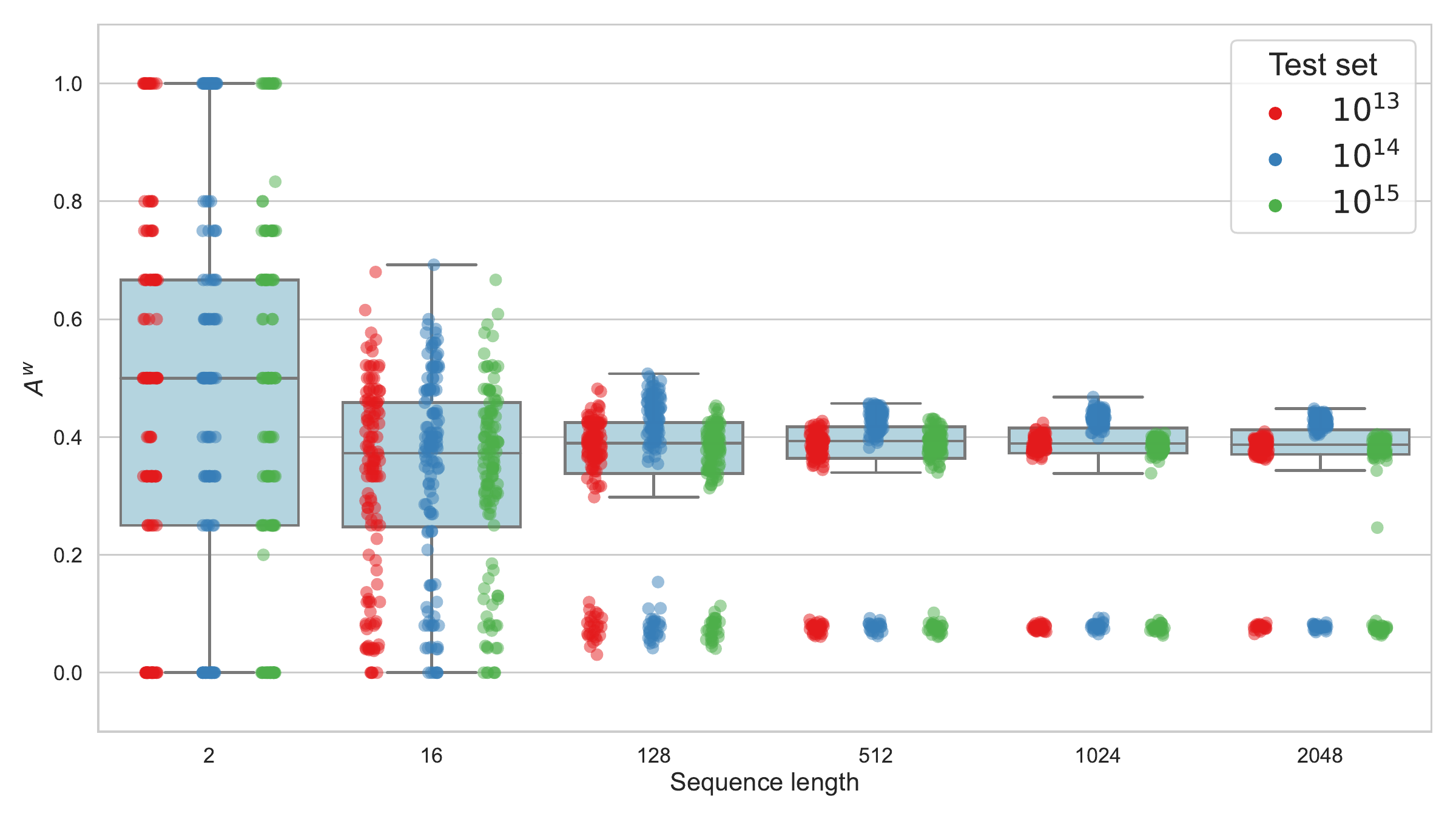}
    \caption{Accuracy obtained from prompts of length $L=2^{10}$, at $\beta=+\infty$, for different generated sequence length $\hat L$. Each box corresponds to a value of $\hat L$ and presents three test sets (marked with different colors) obtained from the first $10^6$ consecutive Dyck words after $10^{13}$, $10^{14}$ and $10^{15}$ (red, blue, green) respectively. Each point represents the result from one out of 128 batches for each of the three test sets. Remarkably, accuracy seems not to be affected by the ``scale'' at which the testing sentence is sampled (i.e., if the test is obtained from sequences around $10^{13}$, $10^{14}$ or $10^{15}$). Results obtained on an architecture with $12$ layers, $12$ heads per layer and $D=2^8$, trained on our dataset of $10^{11}-1$ Dyck words.}
    \label{fig:acc_vs_len}
\end{figure}

\begin{figure}
    \centering
    \includegraphics[width=\linewidth]{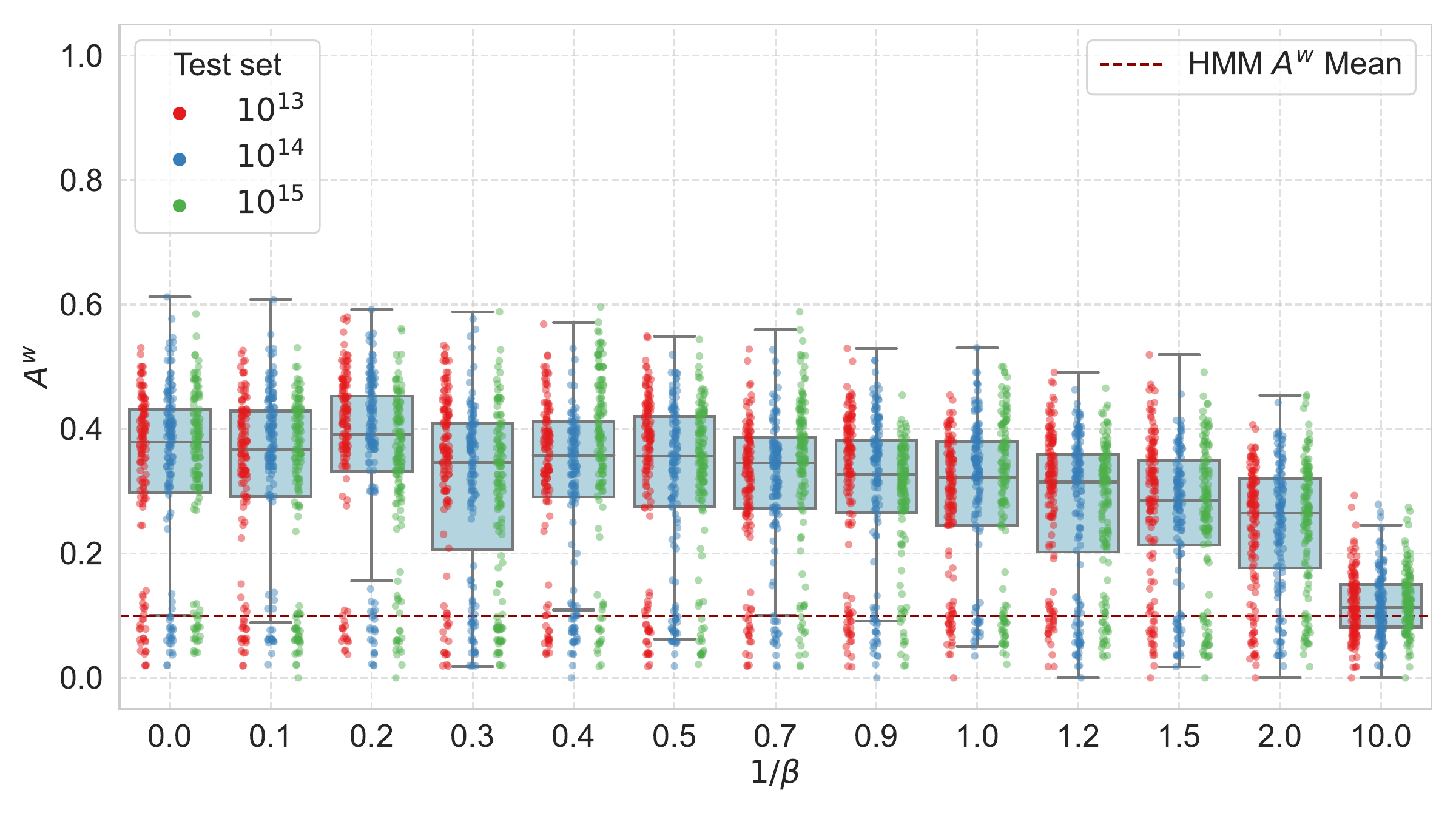}\\
    \includegraphics[width=\linewidth]{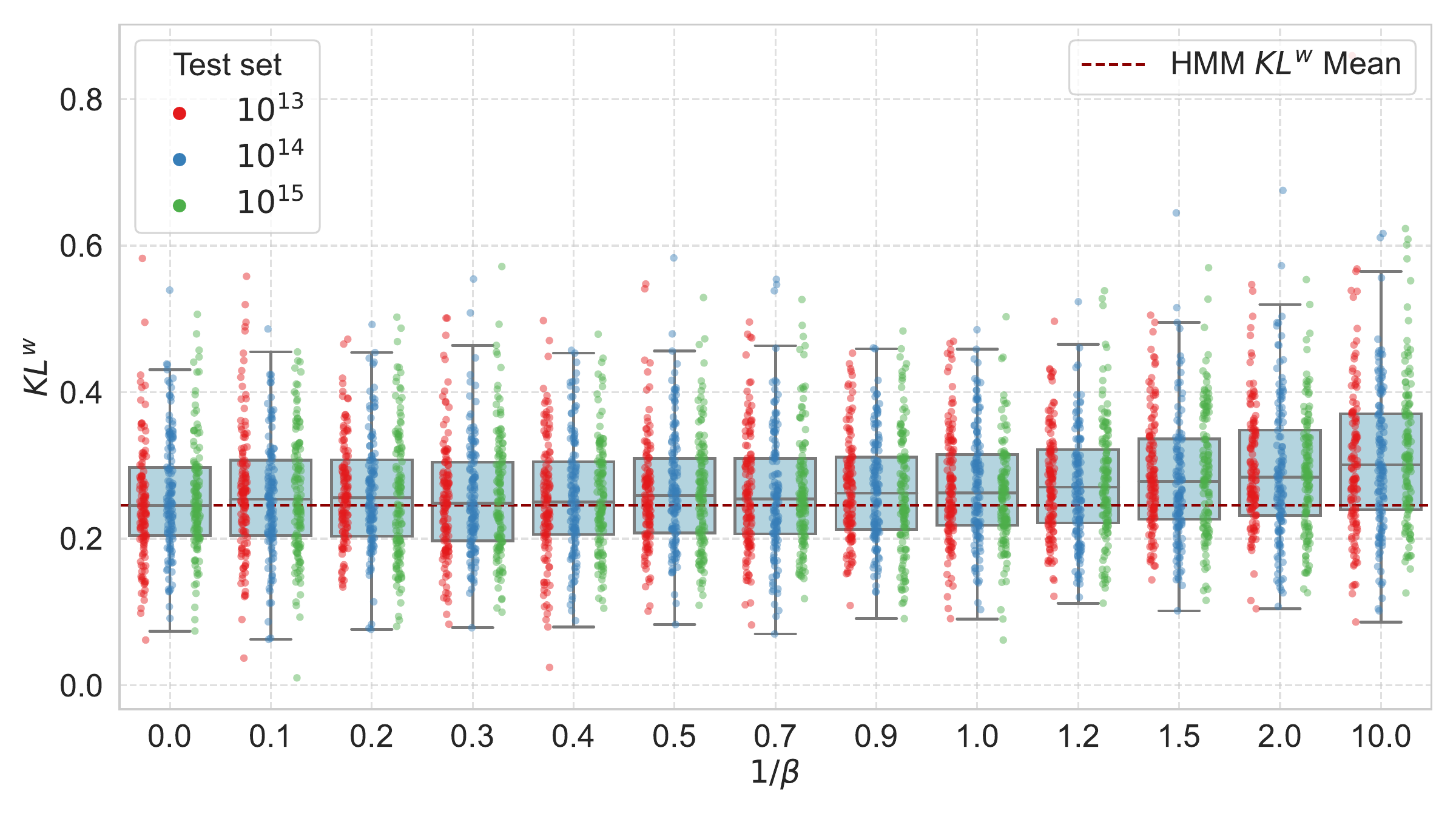}
    \caption{Accuracy for words ($\mathrm{A}^w$, \textit{above}) and Kullback--Leibler divergence for word distribution ($\mathrm{KL}^w$, \textit{below}) are reported at different values of $\beta$, over a set of 128 different inputs of $2^{10}$ tokens. The generated outputs are $\hat L=2^5$ tokens long. The dashed horizontal line represents the HMM performance, whereas boxes correspond to the LLM performance. For each box (corresponding to a given value of $\beta$) test results are given for token sequences sampled from the tokenization of the first $10^6$ words following $10^{13}$ (red), $10^{14}$ (blue) and $10^{15}$ (green), respectively (see also the caption of Fig.~\ref{fig:acc_vs_len}). Despite the fact that the HMM is actually good at reproducing the statistics of the dataset has been trained on, as shown by the divergence $\mathrm{KL}^w$, it fails in predicting the order of Dyck words exhibiting an accuracy comparable to a random guess. Results obtained on an architecture with $12$ layers, $12$ heads per layer and $D=2^8$, trained on our dataset of $10^{11}-1$ Dyck words.}
    \label{fig:acc}
\end{figure}

Figure~\ref{fig:nwp_losses} reports the learning curves for this task, namely the training and validation losses with three different tokenizer vocabulary sizes $D$ ($2^6$, $2^8$, and $2^{10}$ tokens, respectively), rescaled by $\ln D$, value of the loss on the uniform distribution over all the dictionary entries (or, equivalently, the loss in the $\beta\to 0$ case). After an initial rapid decrease, all three curves show an interesting brief initial plateau phase, which anticipates a second descent and appears to extend further in training as $D$ increases: this behavior, to be further investigated in future research, suggests that the model might first initially discover some basic properties of the sequence, while learning only in a second phase more complex relations \cite{rende2024distributional,cagnetta2025scaling}.

In order to measure the performance of the trained model, we feed it a sequence of $L=2^{10}$ tokens from $\mathcal T_{\rm test}$ and produce as output a sequence $\bt=(\bt_a)_{a=1}^{\hat L}$, where in general $\hat L\neq L$. This token sequence is then converted back to the associated Dyck word sequence $\bw(\bt)=(w_a(\bt))_{a=1}^{\hat \ell}$. Due to the tokenization procedure, the value of ${\hat \ell}$ might be smaller or larger than $\hat L$. The obtained sequence is compared  with the true ordered Dyck word sequence $\hat\bw=(\hat w_a)_{a=1}^{{\hat \ell}}$ appearing in $\NT$ after the input sequence fed to the machine. We stress here that it is actually not \textit{a priori} obvious that the token sequence produced by the architecture corresponds to a legit Dyck word sequence. Nevertheless we observed that our LLM almost never produced such illegal Dyck word: for all values of $\beta$ explored (in the range $\beta\geq \sfrac{1}{10}$) the probability (per word) of the appearance of a malformed Dyck word is below $0.25\%$, such value being reached for $\beta=\sfrac{1}{10}$, and no malformed Dyck word was ever observed for $\beta=+\infty$. 

The proficiency of the architecture in NTP strongly depends on the inverse temperature hyperparameter $\beta$ appearing in Eq.~\eqref{eq:soft}, which controls the smoothness of the probability distribution over the tokens: high values of $\beta$ sharpen the probability distribution, by concentrating the probability mass on the most likely outcomes only (with $\beta\rightarrow +\infty$ a deterministic last-layer behavior is obtained); low values instead flatten the distribution, thus increasing the probability of less likely tokens (as $\beta\rightarrow 0$ assigns the same probability to all the tokens effectively producing a random sampler). 

A first metric of performance is the \textit{word accuracy}, simply counting the fraction of guessed words in the generated Dyck word sequence with respect to the true sequence, namely $$
\mathrm{A}^w(\bt)\coloneqq\frac{1}{{\hat \ell}}\sum_{a=1}^{\hat \ell}\mathbb I(w_a(\bt)=\hat w_a).$$ 
We hereby note that a local sequence of undecorated trees, without any information on its position in $\NT$, can present repeated patterns (of variable length) throughout the (infinite) sequence of Dyck words. For this reason, given a sequence of $L$ tokens, it is indeed possible that multiple continuations are allowed and equally correct. Clearly, this phenomenon of multiple plausible continuations gets less likely as the size $L$ of the input prompt grows. In our experiments, with $L=2^{10}$ tokens, we consider as valid only the continuation corresponding to the input picked in the database, thus (slightly) underestimating the performance of the model. Fig.~\ref{fig:acc_vs_len} shows the test accuracy as a function of $\hat L$, evaluated on $128$ input test sequences of length $L$. Each point corresponds to one test sequence. As expected, the accuracy is highest on average when the model is required to predict only a small number of tokens. In this regime, however, there is also greater variability across the test sequences. As the prediction horizon increases, the accuracy gradually stabilizes around $0.4$ and becomes largely independent of the number of tokens to be predicted. Moreover, we clearly see that a solid trend emerges as the length of the generation increases: accuracies are either around $0.4$ (the majority) or just below $0.1$, the value achieved by our markovian baseline, namely the HMM. This may suggest that there are ``hard'' sequences to predict for which the language model is not able to extract any information and, in fact, simply behaves as a markovian model without benefitting from the long context.

A different, useful measure of similarity is $$
\mathrm{KL}^w(\bt)\coloneqq\sum_{\mathclap{w\colon f_w(\bw(\bt))>0}} f_w(\hat\bw) \ln\frac{f_w(\hat\bw)}{f_w(\bw(\bt))}.$$ 
In the expression above, $f_w(\bw)$ is the empirical frequency of the word $w$ in the sentence $\bw$, the sum running on all distinct words with non-zero frequency in $\bw(\bt)$, so that $\mathrm{KL}^w(\bt)$ is the Kullback--Leibler divergence between the empirical frequencies of Dyck words computed from the true and predicted sequences, respectively: this quantity is therefore particularly useful to evaluate if the model is able to capture the statistics of the sequence.

In Fig.~\ref{fig:acc} we report the values of $\overline{\mathrm{A}^w}$ and $\overline{\mathrm{KL}^w}$, the overline denoting the average over the test set of the quantities above, as a function of the hyperparameter $\beta$. The results clearly show that our model largely outperforms the HMM baseline (dashed horizontal line), despite the fact that the HMM exhibits approximately the same Kullback--Leibler divergence of the higher $\beta$ models, and therefore is able to capture the distribution of the tokens as well as our model. The plots also highlight how performance degrades when $\beta<1$, which is not surprising. The best results in terms of word accuracy are obtained at high values of $\beta$, in the range $3.3\leq \beta\leq 10$, while the Kullback--Leibler divergence has its best value for $1.4\leq \beta\leq 3.3$ (very high value of $\beta$ penalize rare tokens). This likely happens because the value of the word accuracy is dominated by high-frequency words, i.e., the correct detection of the words that appear most frequently induces a high value of accuracy.

The aforementioned two quantities do not consider how accurate the model is in predicting rare Dyck words, and in particular prime numbers, which we are clearly very interested in predicting correctly. Specifically, by just accurately predicting the most frequent words in the sequence, a high value for $\overline{\mathrm{A}^w}$ and a low value for $\overline{\mathrm{KL}^w}$ can be obtained. For this reason, we also considered the \textit{precision} $P(w)$, the \textit{recall} $R(w)$, and the \textit{score} $F_1(w)$ of each word. These quantities were computed by feeding $M=10^5$ $L$-tokens-long sentences to the architecture, and observing the predicted sequence. Let $M(w)$ be the number of input sequences that were continued by the word $w$ in the dataset. Let also define $\mathrm{TP}(w)$ the number of \textit{true positives}, namely the number of correctly-guessed appearances of the word $w$; we also call $\mathrm{FP}(w)$ the \textit{false positives}, namely the number of instances in which $w$ was predicted in place of a different word. Finally, let us call $\mathrm{FN}(w)$ the \textit{false negatives}, namely the number of instances in which the word $w$ had to be predicted, but the architecture predicted something else. Then, for each word $w$ having $M(w)>0$ we can define its measured \textit{precision} and \textit{recall} 
\begin{subequations}\label{eq:precision-recall-f1}
\begin{equation}
P(w) \coloneqq \frac{\mathrm{TP}(w)}{\mathrm{TP}(w) + \mathrm{FP}(w)}, \quad R(w)\coloneqq \frac{\mathrm{TP}(w)}{\mathrm{TP}(w)+\mathrm{FN}(w)}=\frac{\mathrm{TP}(w)}{M(w)}, \end{equation}
as well as their harmonic mean, called $F_1$-score, usually denoted by $F_1(w)$
\begin{equation} 
F_1(w) \coloneqq2\left(\frac{1}{P(w)}+\frac{1}{R(w)}\right)^{-1}.
\end{equation}
\end{subequations}
Precision, recall and $F_1$-score for prime numbers are obtained, for example, for $w= \texttt{10}$. We remark that \textit{both} precision \textit{and} recall for a given word $w$ are very important to measure, since a recall equal to 1 could trivially be obtained by always predicting the word $w$, whereas a high precision could be achieved by seldom predicting $w$, e.g., only when the model has a very high confidence. 

\begin{figure}
    \centering
\includegraphics[width=0.95\linewidth]{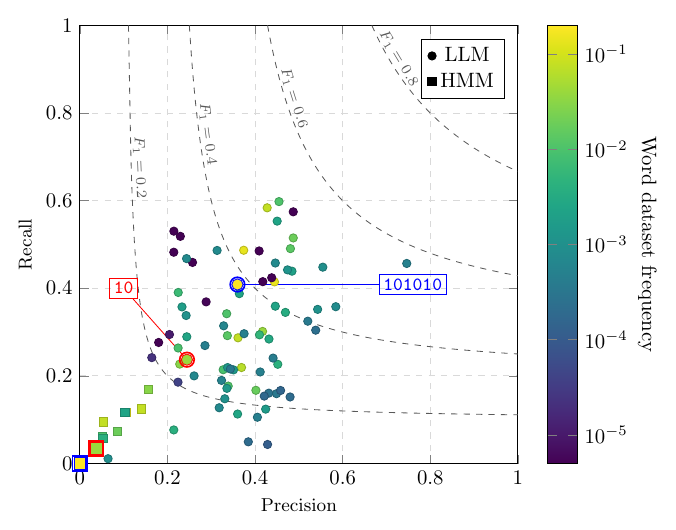}
\caption{Average precision, recall and $F_1$ scores for various Dyck words over $10^5$ test sentences of length $L=2^{10}$ at  $\beta=+\infty$ and dictionary size {$D=2^8$}. Only words appearing at least $50$ times are represented. The plot shows results for the LLM (circles) and the HMM (squares). The LLM clearly outperforms the HMM, whose precision and recall are particularly low. Markers with red frames corresponds to the word \texttt{10}, i.e., to primes, whereas the points in the blue frame correspond to the most frequent Dyck word in the dataset, namely \texttt{101010}.
}
    \label{fig:f-1}
\end{figure}

In Fig.~\ref{fig:f-1} we present the investigation of these quantities by reporting the values of precision, recall, and the $F_1$-score for each word in the dictionary such that $M(w)\geq 50$ in our test set. We can observe that, for example, for prime numbers ($w= \texttt{10}$) the three metrics are all close to $0.3$, indicating that our model correctly detects one prime number roughly every three \textit{existing} primes, as well as correctly predicts one prime number every three \textit{predicted} primes. For other words, such as some square-free integers (e.g., \texttt{1010}, \texttt{101010}, \texttt{10101010}, etc.) the metrics are higher and close to $0.4$--$0.5$.

\begin{figure}
\centering
\includegraphics[width=\linewidth]{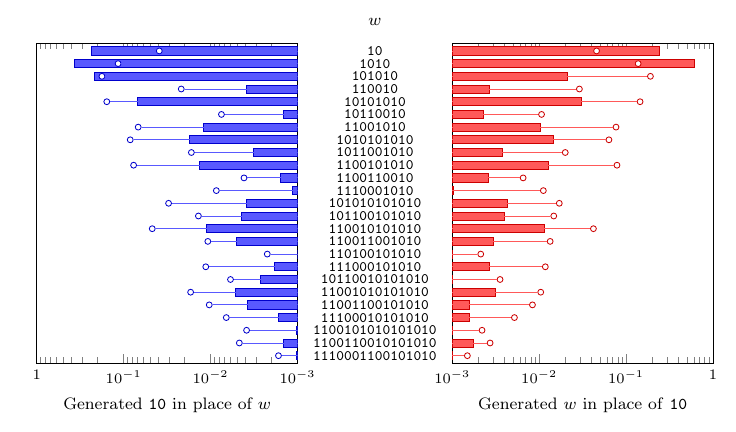}
    \caption{Distributions of the words appearing in the true sequence when the LLM predicts a prime (\textit{left}) and of the predicted words at true prime positions (\textit{right}), obtained via NTP on a set of ${10^5}$ different input prompts of $L=2^{10}$ tokens. The model was trained with a context window of $L=2^{10}$, 12 layers with 12 heads per layer and trained on a dataset covering integers from $2$ till $10^{11}$ with vocabulary size $D=2^8$. For comparison, circles represent the performance of the HMM architecture.}
    \label{fig:primes_prec_rec_gpt}
\end{figure}

\begin{figure}
    \centering
    \includegraphics[width=0.85\linewidth]{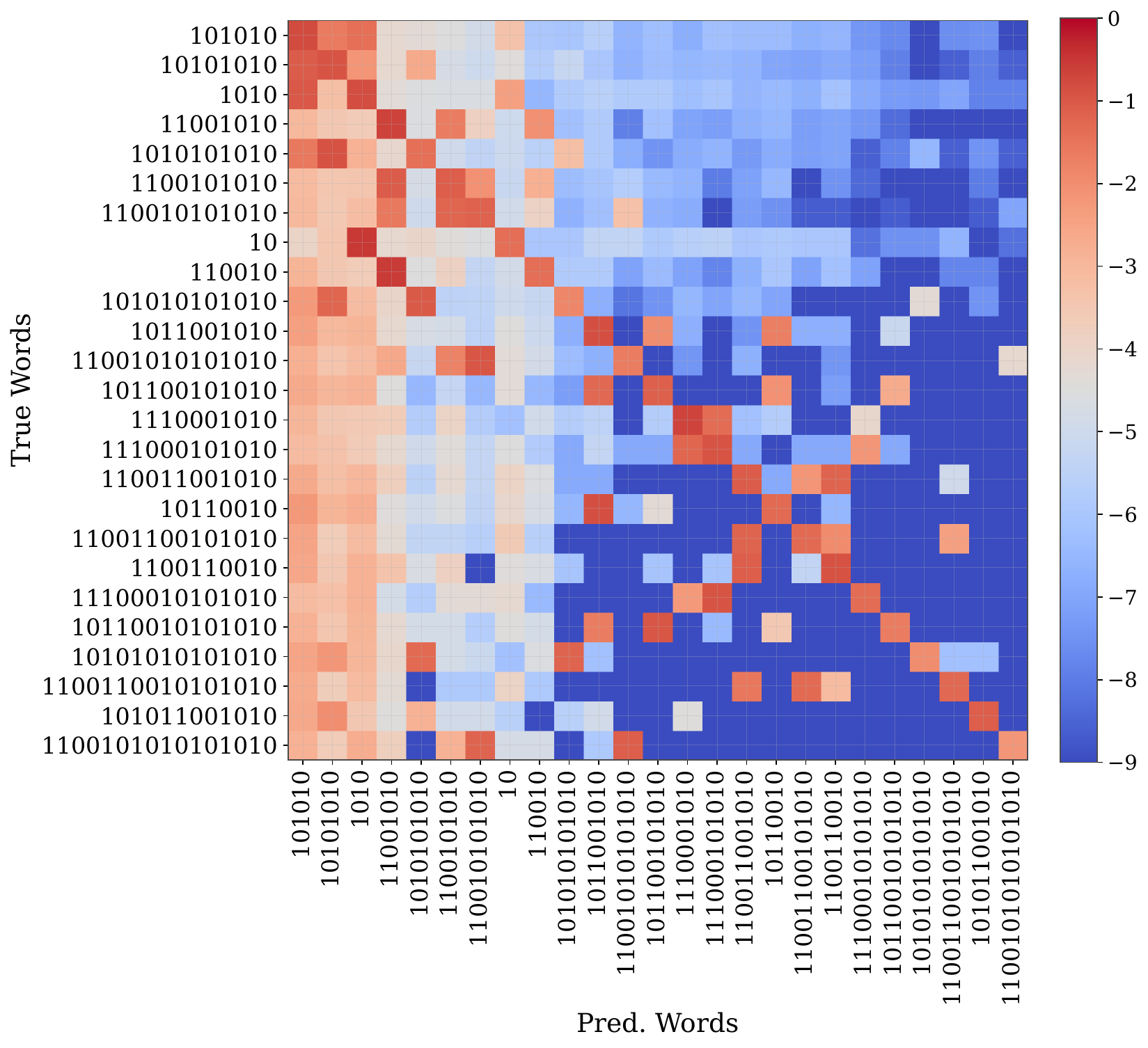}
    \caption{Confusion matrix in the NTP task for the LLM at $\beta=+\infty$. Each row corresponds to a Dyck word and exhibits the logarithm of the frequency at which it is substituted by each one of the Dyck words listed along the $x$-axis. The LLM is trained with a context window of $L=2^{10}$, 12 layers with 12 heads per layer and trained on a dataset covering integers from $2$ till $10^{11}$ with vocabulary size $D=2^8$. Frequencies are obtained by feeding the trained model {$10^5$} sequences of $2^{10}$ consecutive tokens, and querying the model to predict the next tokens until a word is completed. For comparison, the confusion matrix for the HMM is given in \ref{app:hmmconf}.}
    \label{fig:confusion_matrix}
\end{figure}

{In Fig.~\ref{fig:primes_prec_rec_gpt} we present the distributions of words generated by the model in a position occupied by primes in the true sequence and the distribution of words appearing in the true sequence in positions where the LLM predicts a prime. Interestingly, we notice that the model very frequently confuses primes with square-free numbers.} This behavior is confirmed also by the whole confusion matrix, which we report in Fig.~\ref{fig:confusion_matrix}. The confusion matrix $\mathbf C=(c_{ij})$ is such that $c_{ij}$ is the frequency at which the word $i$ is replaced by the word $j$ in the experiment described above for the computation of precision and recall. It is shown that, for example, the sequence \texttt{10} is very often wrongly predicted as \texttt{1010} and, overall, square-free are ``attractors'', as shown by the higher frequency values of the first columns.

\subsubsection{Sensitivity to the $\NT$ Structure:\\Many--Words Correlations and Consecutive Square-Free Integers} 
\begin{figure}
    \centering
    \includegraphics[width=1.\linewidth]{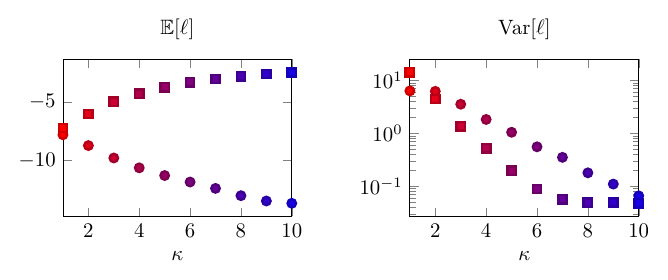}
\caption{Mean (\textit{left}) and variance (\textit{right}) of the log-likelihoods in Eq.~\eqref{eq:LLpt} obtained by sets of $10^5$ synthetic prompts consisting of $2^\kappa$ BPE tokens sampled independently from the empirical marginal token distribution of the training corpus (\textit{circles}) and $10^5$ prompts of equal length from the true sequence of tokens provided by the dataset (\textit{squares}).}
\label{fig:meanv}
\end{figure}

\begin{figure}
\centering
\includegraphics[width=1.\linewidth]{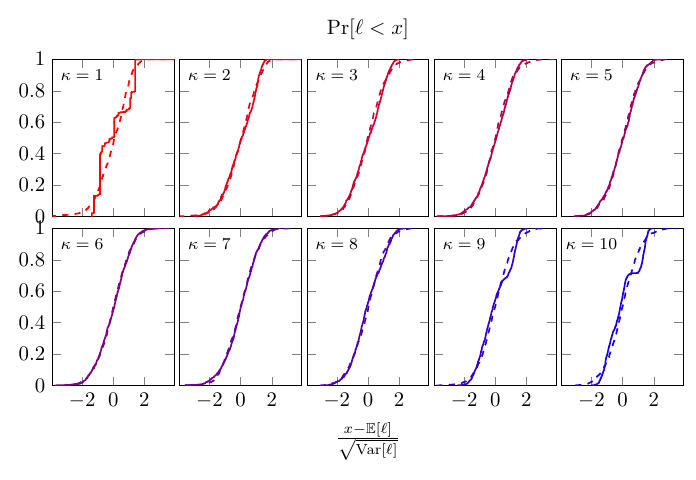}
    \caption{
Cumulative distributions of the standardized variables for various values of $\kappa$, with colors matching the ones in Fig.~\ref{fig:meanv}, obtained from sampling $10^5$ sequences for each $\kappa$. For each $\kappa$, the continuous line refers to the cumulative distribution computed on true sequences, whereas the dashed line of equal color refers to the cumulative distribution computed on plausible, artificially generated sequences, showing that in some cases distributions also markedly differ. The LLM is trained with a context window of $L=2^{10}$, 12 layers with 12 heads per layer and trained on a dataset covering integers from $2$ to $10^{11}$ with vocabulary size $D=2^8$.}
    \label{fig:log-likelihood-flipped}
\end{figure}

To assess whether the model captures sequential dependencies beyond marginal token statistics, we compared the log-likelihoods assigned to two sets of prompts. The first set consists of authentic BPE-token sequences extracted from the validation corpus. The second consists of synthetic prompts generated by independently sampling BPE tokens from the empirical marginal distribution estimated on the tokenized training corpus. By construction, the synthetic generation process preserves the marginal BPE-token probabilities in expectation but does not preserve the empirical counts exactly within each sampled prompt\footnote{Note that, because a BPE token may span multiple Dyck words or represent only part of one, matching the marginal distribution of BPE tokens does not imply matching the frequencies of complete Dyck words.}. Independent sampling also removes the higher-order dependencies between successive tokens present in the original arithmetic sequence.
For each tokenized sentence $\bt=(\bt_a)_{a=1}^L$ of $L>1$ BPE tokens, we computed the average conditional log-likelihood per token:
\begin{equation}\ell(\bt)=\frac{1}{L-1}\sum_{k=1}^{L-1}\ln p_{\btheta}(\bt_{k+1}|\bt_{1:k}),\label{eq:LLpt}\end{equation}
the conditional probability being the one provided by the trained model. The results, shown in Fig.~\ref{fig:meanv} and Fig.~\ref{fig:log-likelihood-flipped}, are reported as a function of $\kappa$, $L=2^\kappa$ being the length of the evaluated sequence. Results are obtained from a set of $10^5$ prompts from the validation set and $10^5$ synthetic prompts constructed by independently sampling BPE tokens from the empirical marginal token distribution. As $\kappa$ increases, the mean of the distribution of the log-likelihoods of the artificial prompts markedly diverges from the corresponding value obtained computing the same distribution on true sequences, whereas the shape is particularly affected in the case of short and long sequences. The model therefore does not evaluate a sequence through its prior token probabilities, finding sequences generated from the empirical unigram token distribution without preserving sequential dependencies very unlikely, and the amount of evidence it extracts increases with the available context.

As a second test, we investigated whether GPT-2 has learned a non-trivial structural property of square-free integers, that is, four or more consecutive square-free positive integers are not allowed in the sequence. Equivalently, every sequence of consecutive square-free integers has length at most three.
\begin{figure}
    \centering
\includegraphics[width=\linewidth]{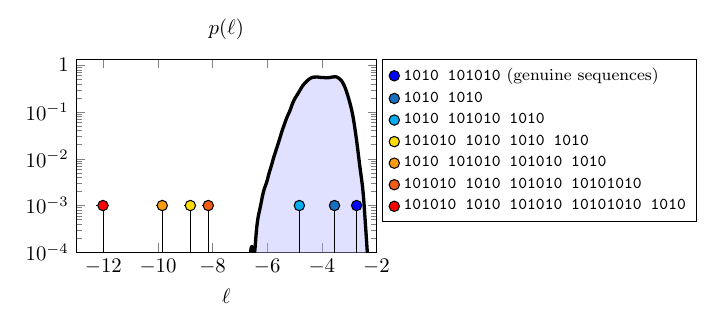}
\caption{Distribution of the log-likelihood $\ell$ obtained via kernel density estimation from $10^5$ true sequences from the test set consisting of $20$ \textit{Dyck words}. Dots denote the average values of the likelihood obtained by sampling from the dataset $2^8$ sentences ending in \texttt{1010 101010}: such ending is substituted with sequences of square-free Dyck words, the log-likelihood of the unaltered sequences being marked in pure blue. It is shown that the modified sequences representing more than three consecutive square-free integers (reddish dots) are placed at very low values of $\ell$, whereas true sequences and fake sequences with no more than three consecutive square-free (bluish dots) are assigned a higher log-likelihood.}
    \label{fig:square_free}
\end{figure}
Fig.~\ref{fig:square_free} suggests that GPT-2 has in fact ``learned'' this structural property. As in the previous experiment, we computed the frequency distribution of the log-likelihood in Eq.~\eqref{eq:LLpt} when computed on test set sequences. We sampled $2^8$ sequences ending with \texttt{1010 101010}, measured their average log-likelihood, and then constructed 
a set of artificial prompts by replacing the ending \texttt{1010 101010} with sequences of consecutive square-free integers. As shown in Fig.~\ref{fig:square_free}, the insertion of representations of up to three consecutive square-free integers results in log-likelihood values within the real sequence distribution support. In contrast, prompts in which more than three consecutive square-free integers are inserted are systematically associated to substantially lower log-likelihoods. Since such sequences are mathematically impossible, this result is consistent with the model learning a local statistical signature of the constraint.

\subsubsection{Scalings and hyperparameters}

To gain a deeper understanding of the learning dynamics of our language model, we performed a systematic scaling study, investigating how predictive performance depends on four fundamental factors: the size of the training dataset $n$, the model size, the context length $L$, and the tokenizer vocabulary size $D$. While empirical scaling laws have been extensively characterized for natural language corpora, it is not obvious that they extend to synthetic datasets such as ours. In contrast to natural language, the corpus $\NT$ is generated by deterministic mathematical rules and exhibits highly structured dependencies. This raises the question of whether the scaling behavior of large language models trained on such data follows the same trends observed in conventional language modeling. For each scaling experiment, we measured the model accuracy after training as a function of the parameter under investigation. 

\begin{figure}[!tb]
    \centering
    \includegraphics[width=1.\linewidth]{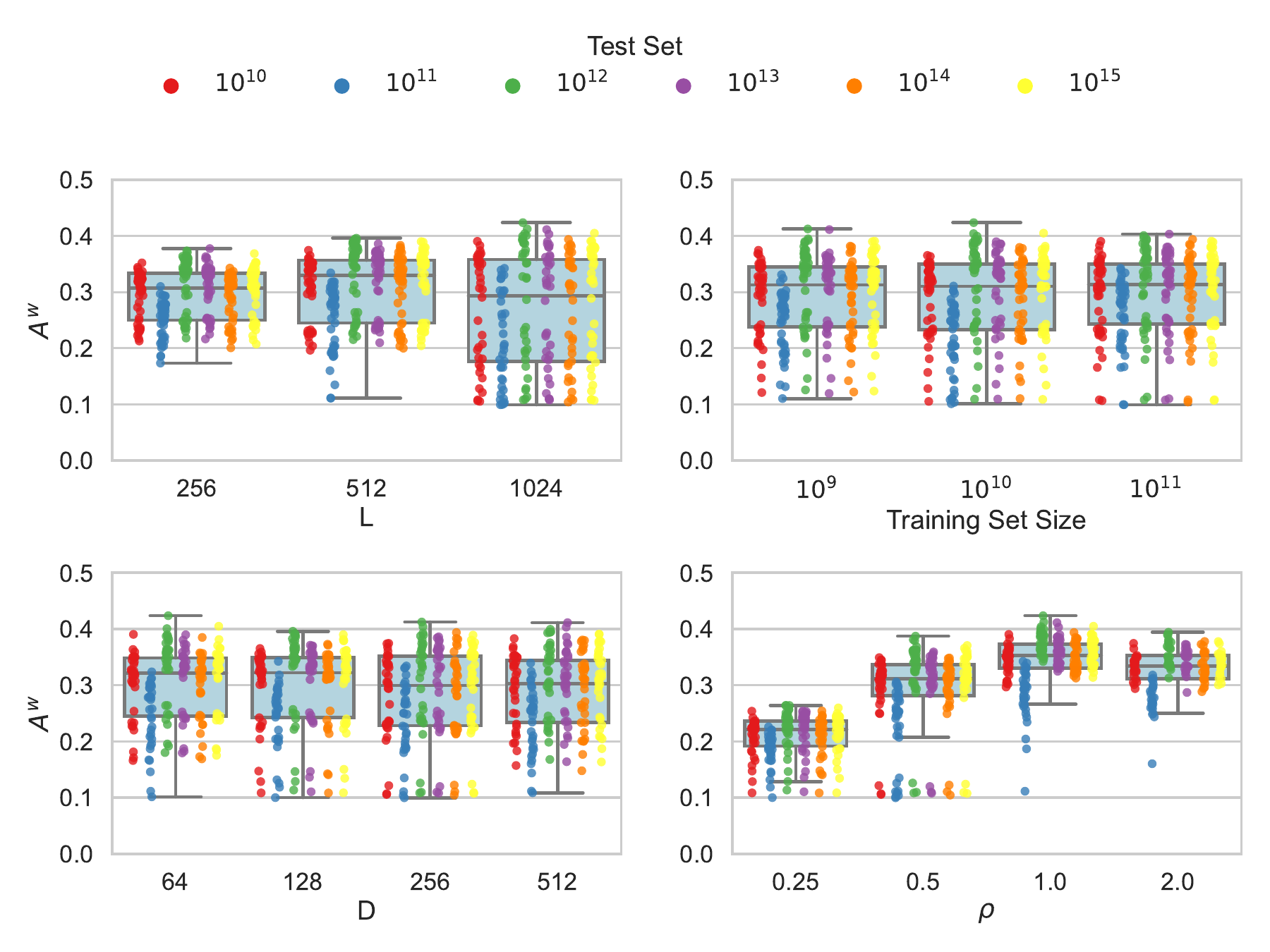}
    \caption{Accuracy at $\beta=+\infty$ for various combinations of the hyperparameters considered in the experiments. The reference model is described in the text and corresponds to scale $\rho=1$, it is trained on the dataset of the first $n=10^{11}$ Dyck words with context window size $L=2^{10}$ and vocabulary size $D=2^8$. In the panels, one of such parameters is altered by keeping the remaining parameters fixed: from left to right, $L$, $n$, $D$ and $\rho$. Each dot represents the accuracy of a single architecture trained with the indicated hyperparameters on a sentence of length $L$ at a test scale ranging from $10^{10}$ to $10^{15}$, each scale being highlighted by a different color. Each of the four plots shows the effect of each hyperparameter, obtained grouping all the set of accuracy values only by this latter. }
    \label{fig:acc_boxplots}
\end{figure}

Unless otherwise specified, accuracy was computed by averaging over a batch of $128$ sequences, each with a prediction horizon equal to the context length of the corresponding model. To ensure a fair comparison across different model configurations, each experiment was preceded by a learning-rate selection phase, by following the same protocol as described above in the \textit{Training} paragraph of Section~\ref{sec:experiments}. We adopted as reference model the architecture described in Section~\ref{sec:model}, namely an OpenAI GPT-2-type architecture \cite{radford2019language} with $12$ layers, each combining sequentially $12$ heads each of dimension $64$, followed by a layer normalization and a feed-forward neural network using the GeLU activation function, resulting in an architecture with $P=8.7\cdot 10^7$ trainable parameters. We scaled this architecture by a scaling factor $\rho$, so that a model of scale $\rho$ with respect to the reference one has $12\rho$ layers each with $12\rho$ attention heads, resulting in an embedding dimension of $d=768\rho$ of the multi-head attention layer. We explored the performance of the model for $\rho=\sfrac{1}{4}$, $\rho=\sfrac{1}{2}$, $\rho=1$ and $\rho=2$, keeping $D=2^{8}$ and $L=2^{10}$ with a dataset of $n=10^{11}$. Accuracy results are reported in the rightmost panel in Fig.~\ref{fig:acc_boxplots}. There, each point corresponds to the accuracy of the architecture on a sentence of length $L$, different colors corresponding to testing at different scales on unseen sentences at scales ranging from $10^{10}$ (within the training set range) to $10^{15}$ (much beyond the training set range). Test accuracy seems to increase with $\rho$ up to $1$, with results obtained by the model with $\rho=2$ being comparable with (if not worse than) the ones achieved by our reference model: if the increase in accuracy with $\rho$ reflects the greater expressivity of the model, the saturation at $\rho=1$ might be a sign of the fact that the larger model would require a larger dataset to express its power. This experiment has been a partial empirical validation of the adopted model size.

By fixing the architecture, we also changed (one at a time) the hyperparameters $D$ (dictionary size), $L$ (context window) and the training set size $n$, testing the obtained trained model on sentences sampled at different scales in unseen regions of the sequence, ranging from order $10^{9}$ to order $10^{15}$. In all cases accuracy is very mildly affected by the hyperparameters variations, only exhibiting larger fluctuations when $L$ is large.

\subsection{Masked Language Modelling}

\begin{figure}[!tb]
    \centering
    \includegraphics[width=\linewidth]{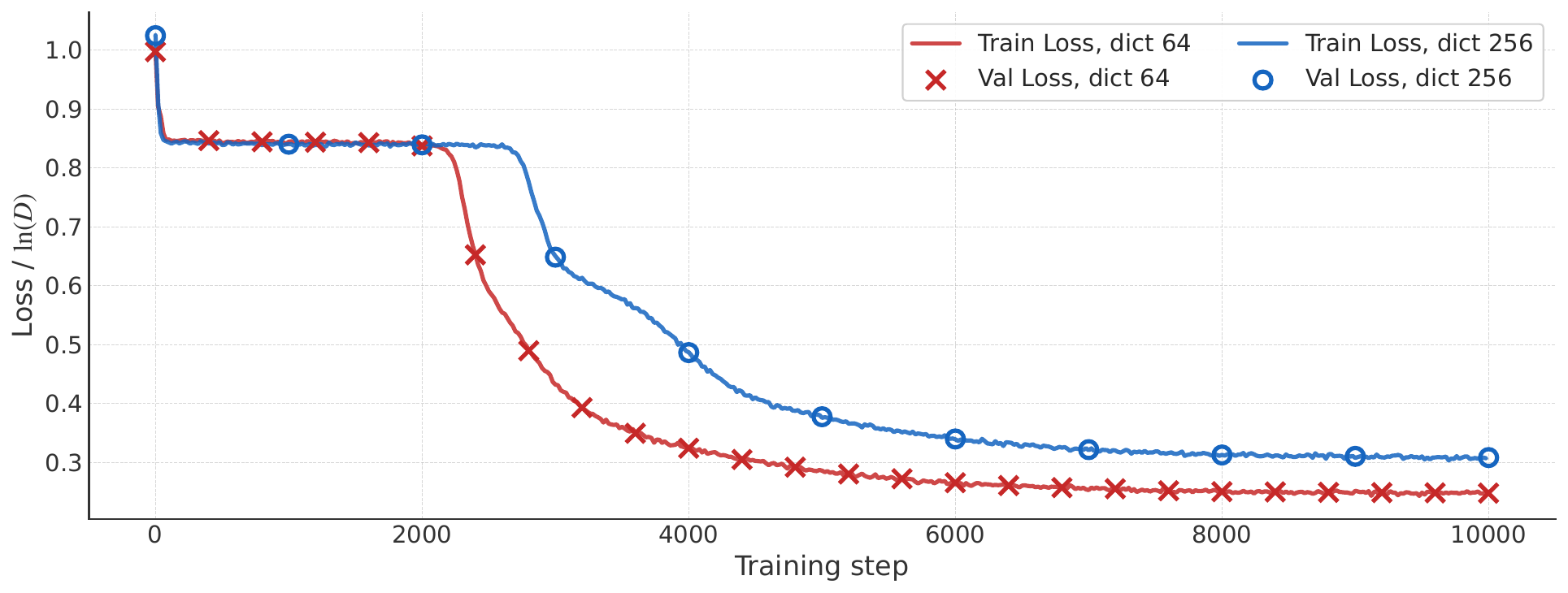}
    \caption{Loss curves for an architecture trained via masked language modelling. Here, tokenizer vocabulary sizes equal to $D=2^6$ and $D=2^8$. The loss value is rescaled by $\ln D$, loss value associated to a uniform distribution.}
    \label{fig:losses_masked}
\end{figure}

Let us now briefly discuss the results obtained for the second considered training task, namely masked-token prediction. As anticipated, here the goal is to predict missing tokens within a sequence that have been \textit{masked} in the test sequences, i.e., removed at inference time. The training is performed by masking $15\%$ of the tokens in each sequence of $L=2^{10}$ tokens, as described above. A crucial, additional parameter of the experiment is the percentage of tokens that we mask in our test sequences, which we name $p_m$ and does not coincide in general with the probability of observing a masked token during training.
Similarly to the NTP task case, in Fig.~\ref{fig:losses_masked} we report the learning curves for two different vocabulary sizes ($D=2^6$ and $D=2^8$ tokens, respectively). In this task, the initial plateau phase appearing in the NTP case is even more evident and extends further as $D$ increases.

\begin{figure}[!tb]
    \centering
    \includegraphics[width=0.85\linewidth]{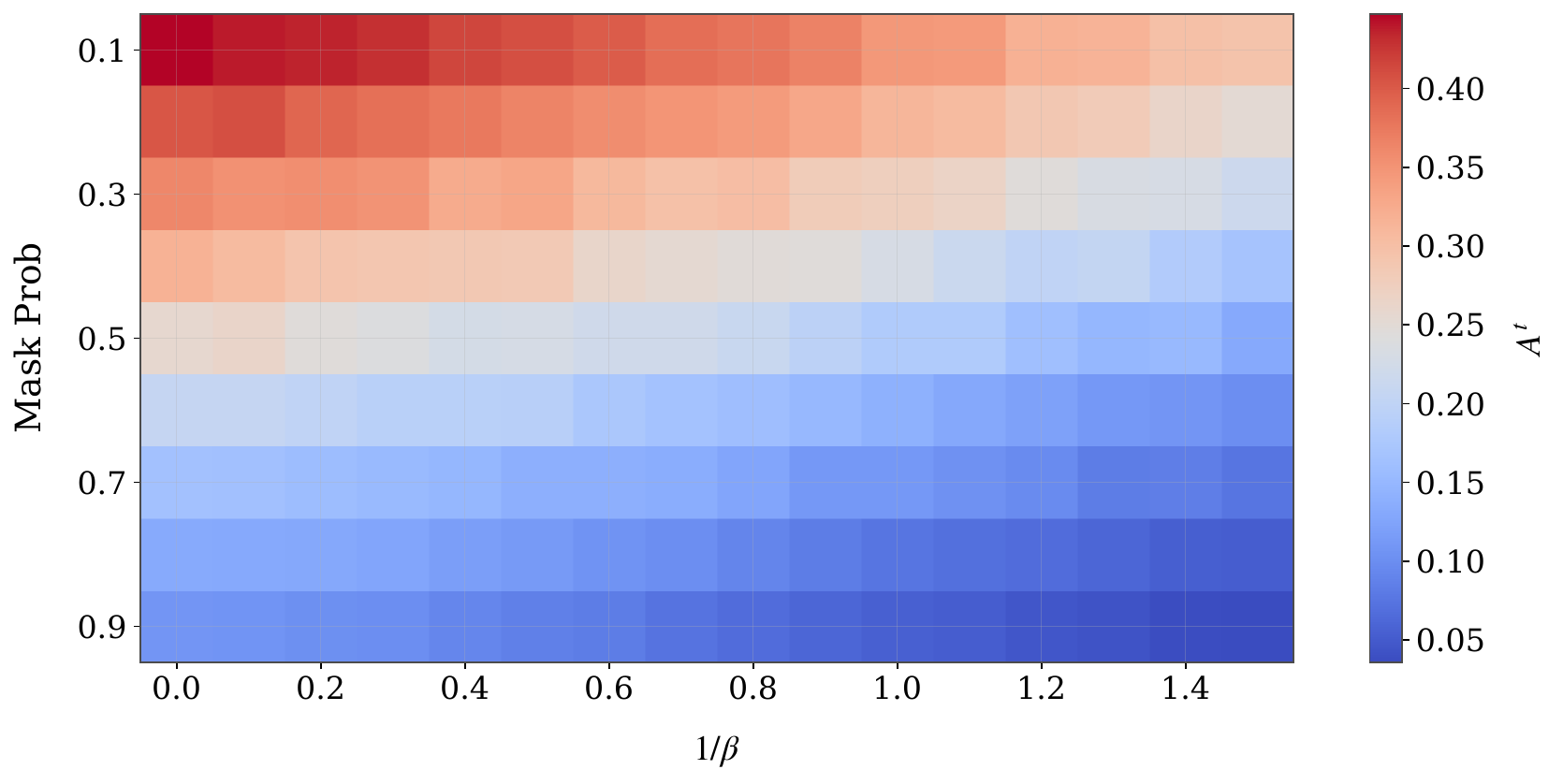}
    \caption{Token accuracy as a function of masking probability and $\beta$ within the Masked Language Modelling strategy on the test set.}
    \label{fig:acc_vs_pm_masked}
\end{figure}

As in the NTP case, the output probability distribution is given in the form in Eq.~\eqref{eq:soft} and depends therefore on the hyperparameter $\beta$. We analyzed the joint impact of the inverse temperature $\beta$ and the masking probability $p_m$ in Fig.~\ref{fig:acc_vs_pm_masked}. Here we plot the performance of the model by adopting as a metric the \textit{token accuracy},
\[\mathrm{A}^t(\bt)\coloneqq\frac{1}{|\mathcal J(\bt)|}\sum_{a\in\mathcal J(\bt)}\mathbb I(\bt_a=\hat\bt_a),\]
namely an accuracy measure on the token sequence that compares a generated sequence $\bt=(\bt_a)_{a=1}^L$ with the corresponding true token sequence $\hat\bt=(\hat\bt_a)_{a=1}^L$ obtained by tokenizing the ground truth word succession. A clear pattern appears indicating that performance degrades when both $p_m$ and $\beta^{-1}$ grow, which is of course not surprising. Within the high-$\beta$ regime (in the range $3.3\leq \beta\leq 10$), the accuracy on tokens is above $0.4$, remaining above $0.3$ even with values of $p_m$ and $\beta^{-1}$ growing up to $0.4$ and $0.5$, respectively.

\begin{figure}[!tb]
    \centering
    \includegraphics[width=\linewidth]{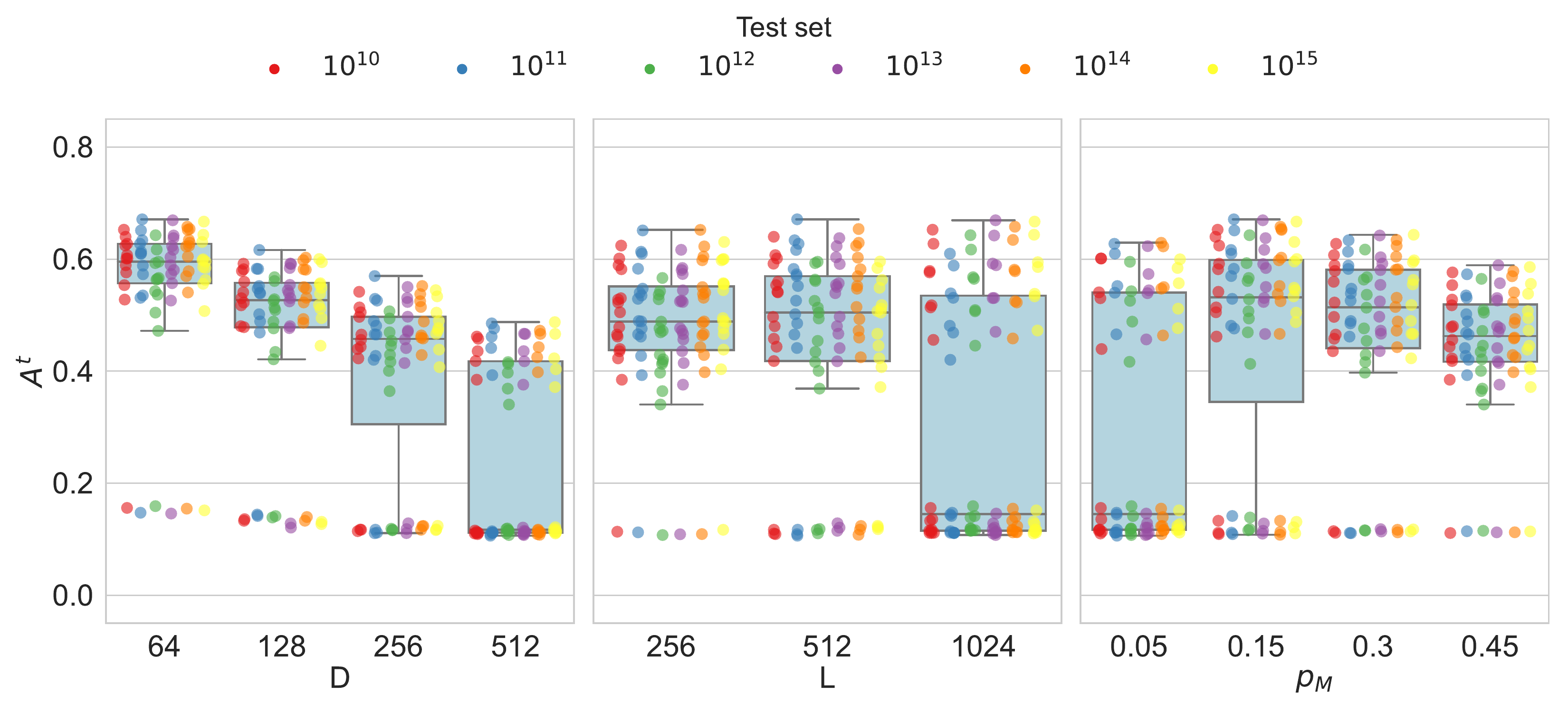}
    \caption{Accuracy at $\beta=+\infty$ for various combinations of the hyperparameters considered in the MLM experiments. The model size correspond to $\rho=1$ scale in the experiments above, it is trained on the dataset of the first $n=10^{11}$ Dyck words. In the panels, one parameter is altered in turn, by keeping the remaining ones fixed: from left to right, we tuned the tokenizer dictionary size $D$, the context window $L$ and the masking probability at training time $p_M$. Each dot represents the accuracy of a batch of $2^7$ sequences, with the architecture trained with the indicated hyperparameters on a sentence of length $L$ at a test scale ranging from $10^{10}$ to $10^{15}$, each scale being highlighted by a different color. Each of the three plots shows the effect of each hyperparameter, obtained grouping all the set of accuracy values only by this latter.
    The results highlight the detrimental effect of a large $D$; as $D$ increases, tokens often encompass multiple words, creating a harder masking task during training. Sequence length $L$ does not significantly impact the training procedure, aside from requiring more refined learning rate tuning at $L=2^{10}$. Optimal performance is achieved at a masking probability of $p_m=0.15$  }
    \label{fig:sweep_mlm}
\end{figure}

Finally, in Fig.~\ref{fig:sweep_mlm} we present the token accuracy in the MLM task as a function of the $D$, $L$ and $p_m$ on our test set. We see a mild dependence on these parameters, except for $D$: larger values of $D$ are associated in fact to lower accuracy levels. As in the NTP task, we observe a marked separation between a set of ``easy sequences'', which are completed with high accuracy, and ``hard sequences'', for which the model relies on a random guess.

\section{Conclusions and Future Work}\label{sec:conclusions}

We trained a GPT-2 architecture from scratch on the Dyck-word representation
of the first $10^{11}$ integers and measured its predictive ability on the
sequence $\NT$. At the optimal values of the inverse temperature $\beta$, the
word accuracy in Next-Token prediction settles around $0.4$, well above the
Hidden Markov Model baseline, which reproduces the marginal statistics of the words
but fails to predict their order. At the level of individual words, precision,
recall and $F_1$ are close to $0.3$ for primes and reach $0.4$--$0.5$ for
several square-free words; the dominant error mode is a systematic confusion
between square-free integers including primes, consistent with the proximity of the
corresponding Dyck words (different concatenations of \texttt{10}).
The accuracy remains stable around the same value on test blocks located at
$10^{13}$, $10^{14}$ and $10^{15}$: since the sequence of trees is not
translationally invariant, this stability far beyond the training interval
is a non-trivial form of out-of-distribution generalization.

Two further tests indicate that what the model has captured is not merely a
frequency profile. Comparing the log-likelihood of authentic prompts with that
of synthetic prompts generated by independently sampling BPE tokens from the empirical marginal token distribution of the tokenized training corpus, the two distributions overlap for very short
prompts and separate progressively as the evaluated context grows, becoming disjoint beyond a few tens of BPE tokens. The likelihood therefore depends on sequential dependencies between BPE tokens rather than solely on their marginal frequencies. Second, the model assigns
sharply lower likelihood to sequences containing more than three consecutive
square-free integers, a configuration that arithmetic forbids, whereas the
Hidden Markov baseline places such sequences inside the bulk of the true
prompts. We record this as an empirical result: the measurement establishes
that the constraint leaves a statistical signature in the likelihood assigned
by the model, not that the model has represented the arithmetic fact that
produces it.

The scaling experiments qualify these results. Within the ranges explored,
the accuracy is essentially insensitive to the context window, the dataset
size and the vocabulary size, while it depends on the model size, with the
standard GPT-2 configuration performing best. This insensitivity concerns the
accuracy of the generated continuation, and coexists with the context
dependence visible in Fig.~\ref{fig:log-likelihood-flipped}: the ability to
discriminate an authentic sequence from an order-destroyed one improves
steadily with the length of the evaluated context, while the ability to
reproduce the correct continuation does not.
{\color{black} The contrast deserves emphasis, because the corpus is known to
carry genuine long-range structure: the correlation analysis of
\cite{Modena2025} shows a diffusive-to-superdiffusive crossover in the walks
associated with individual Dyck words, over lags far exceeding any window used
here. Prediction and discrimination are thus limited by different features of
the corpus, and only the second appears to benefit from that structure.}

Some limitations should be kept in mind. The experiments concern a single
canonical trajectory --- there is only one $\NT$ --- so the usual averaging
over independent samples of the data-generating process is not available;
variability was assessed across disjoint blocks of the sequence instead.
{This substitution rests on the approximate homogeneity
quantified in Fig.~\ref{fig:jensen}, and its validity is limited by the same
measurement: the corpus admits no homogeneous bulk, and what the stability of
the accuracy at $10^{13}$--$10^{15}$ shows is that the trained model inherits
that approximate homogeneity rather than that a translation-invariant limit
exists. Identifying the notion of limit appropriate to a corpus of this kind
is, in our view, the question that the present measurements leave open.}

Future work will focus on probing the internal representations of the trained
networks, to assess whether the geometry of the embedding space reflects
algebraic proximity among integers, along the lines of structural probes
developed for natural language \cite{hewitt2019structural,diego2024polar,
caucheteux2021disentangling,acevedo2025differential}. The role of the tokenization scheme, which
was kept fixed here, also deserves a systematic comparison. Downstream tasks, and
in particular the comparison with existing factorization algorithms, belong
to the same program and are left for subsequent investigation.

\section*{Acknowledgments}

The authors are grateful to Alina S\^{i}rbu for providing the code to generate the database and Riccardo Rende for preliminary code and experiments on MLM. This research was performed under the auspices of Italian National Group of Mathematical Physics (GNFM) of the National Institute for Advanced Mathematics - INdAM.
The authors acknowledge the financial support from the European Union - Next Generation EU - Grant PRIN 2022B5LF52. This project received support from the EU H2020 ICT48 project Humane AI Net (grant no. 952026), the Italian Ministry of University and Research PRIN 2022 (code J53D23003690006), and the Italian Extended Partnership PE01—FAIR (Future Artificial Intelligence Research, proposal code PE00000013) under the National Recovery and Resilience Plan. Federica Gerace is supported by European Union-NextGenerationEU (NGEU) and she is partially supported by project SERICS (PE00000014) under the  MUR National Recovery and Resilience Plan.

\printbibliography
\newpage
\appendix
\section{Details on the training process}\label{app:training}
In this appendix, we detail the experimental setup, computational resources, and training procedures used to develop and evaluate our models. All experiments were conducted using four NVIDA L40S GPUs. To ensure the robustness of our findings, several random seeds have been tested, showing no relevant differences in the final results. 
In Fig.~\ref{fig:nwp_losses} and Fig.~\ref{fig:losses_masked} we have represented the loss value as a function of the \textit{training step}: during a single training step, the model is fed a batch of $2^6$ sequences of $2^{10}$ tokens randomly selected from the training set; the resulting loss is then used for a backpropagation step updating the weights via gradient descent. The selection procedure allows for overlaps between sentences in different batches. 
As a result, a default training run processed approximately $9\cdot 10^8$ tokens. To prevent overfitting over the course of the updates, we implemented an early stopping mechanism. For all subsequent inference and generation tasks, we utilized the standard token sampling procedures provided by the \texttt{PyTorch} framework. Code and adopted datasets are available upon request. 

\section{Confusion matrix for the Hidden Markov Model}
\label{app:hmmconf}
\begin{figure}
\includegraphics[width=0.9\linewidth]{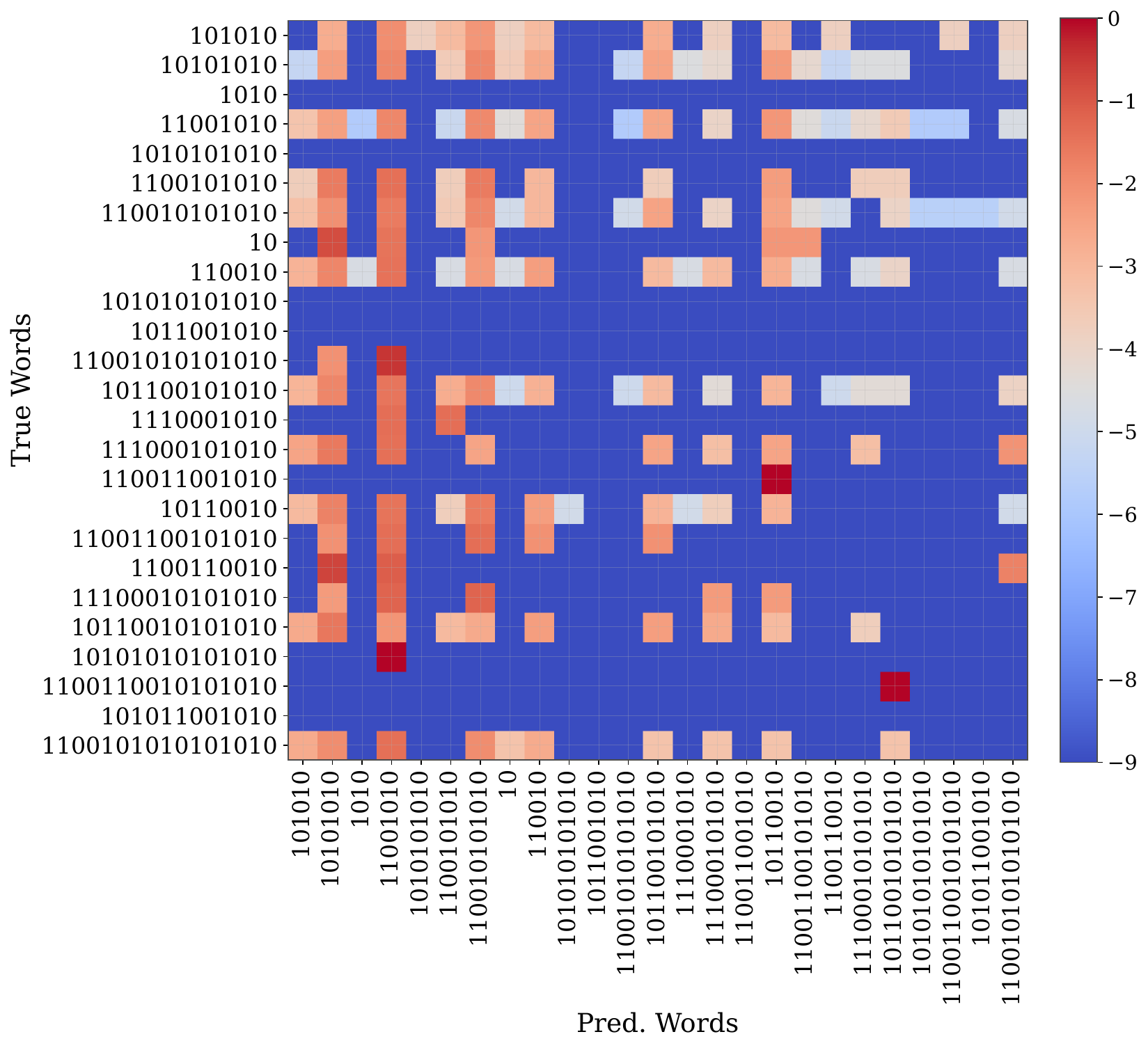}
\caption{Confusion matrix for the trained HMM. It is reported the logarithm of the frequency at which a true word appears as replaced by a different word in the dictionary in the HMM generative process.}
\label{fig:confhmm}
\end{figure}

In Fig.~\ref{fig:confhmm} we report the confusion matrix for the HMM, to be read analogously to the LLM confusion matrix in Fig.~\ref{fig:confusion_matrix}.
Unlike the LLM case, the matrix, representing the logarithm of the frequency at which a given word is replaced by another in the model generation, does not exhibit a structure and, in particular, it can be observed that the diagonal (corresponding to the occurrence a correct prediction) is not particularly highlighted. The markovian nature of the model, in fact, does not allow the information from the preceding sequence to be used for the prediction.
\end{document}